\documentclass[acmsmall]{acmart}
\usepackage{rotating}
\usepackage{amsmath}
\usepackage{amsfonts}
\usepackage{color}
\usepackage{graphicx}

\usepackage{amssymb}
\usepackage{subfigure}
\usepackage{verbatim}
\usepackage{algorithmicx,algorithm}
\usepackage[noend]{algpseudocode}
\usepackage{makecell}
 \usepackage{multirow}
\AtBeginDocument{%
  }

\setcopyright{acmcopyright}
\copyrightyear{2022}
\acmYear{2022}
\acmDOI{XXXXXXX.XXXXXXX}

\acmJournal{JACM}
\acmVolume{37}
\acmNumber{4}
\acmArticle{111}
\acmMonth{8}

\begin{document}

\title{Multimodal Dialog Systems with Dual Knowledge-enhanced Generative Pretrained Language Model}

\author{Xiaolin Chen}
\email{e-mail: cxlicd@gmail.com}
\affiliation{%
  \institution{School of Software, Joint SDU-NTU Centre for Artificial Intelligence Research, Shandong University}
  \country{China}
  }

\author{Xuemeng Song$^*$}
\affiliation{
  \institution{School of Computer Science and Technology, Shandong University}
  \country{China}}
\email{sxmustc@gmail.com}

\author{Liqiang Jing}
\affiliation{%
  \institution{School of Computer Science and Technology, Shandong University}
  \country{China}}
  % \city{Jinan}
  % \state{Shandong}
  % \country{China}
  % \postcode{250101}}
 \email{jingliqiang6@gmail.com}

\author{Shuo Li}
\affiliation{%
  \institution{School of Computer Science and Technology, Shandong University}
  \country{China}}
  % \city{Jinan}
  % \state{Shandong}
  % \country{China}
  % \postcode{250101}}
 \email{zile020401@gmail.com}

\author{Linmei Hu}
\affiliation{%
  \institution{School of Computer Science and Technology, Beijing Institute of Technology}
  \country{China}}
%   \city{Beijing}
% %   \state{Shandong}
%   \country{China}
%   \postcode{100081}}
 \email{hulinmei@bit.edu.cn}

% the School of Computer Science and Technology, Beijing Institute of Technology, Beijing 100081, China. (e-mail: hulinmei@bit.edu.cn
 
% \author{Linmei Hu}
% \affiliation{%
%   \institution{Beijing University of Posts and Telecommunications}
%   \city{Beijing}
% %   \state{Shandong}
%   \country{China}
%   \postcode{100876}}
%  \email{hulinmei@bupt.edu.cn}

\author{Liqiang Nie$^*$}
\affiliation{%
  \institution{School of Computer Science and Technology, Harbin Institute of Technology (Shenzhen)}
  \country{China}}
  % \city{Shenzhen}
  % \state{Guangdong}
  % \country{China}
  % \postcode{518055}}
 \email{nieliqiang@gmail.com}
\thanks{$^*$Corresponding authors: Xuemeng Song and Liqiang Nie.}

\renewcommand{\shortauthors}{Chen et al.}

\begin{abstract}
  Text response generation for multimodal task-oriented dialog systems, which aims to generate the proper text response given the multimodal context, is an essential yet challenging task.
Although existing efforts have achieved compelling success, they still suffer from two pivotal limitations: 1) \emph{overlook the benefit of generative pre-training}, and 2) \emph{ignore the textual context related knowledge}.
To address these limitations, we propose a novel dual
knowledge-enhanced generative pretrained language model
for multimodal task-oriented dialog systems~(DKMD),
consisting of three key components: \emph{dual knowledge selection}, \emph{dual knowledge-enhanced  context learning}, and \emph{knowledge-enhanced response generation}.
To be specific, the dual knowledge selection component aims to select 
the related knowledge according to both textual and visual modalities of the given context.
Thereafter, the dual knowledge-enhanced context learning component targets seamlessly integrating the selected knowledge into the multimodal context learning from both global and local perspectives, where the cross-modal semantic relation is also explored.
Moreover, the knowledge-enhanced response generation component comprises a revised BART decoder, where an additional dot-product knowledge-decoder attention sub-layer is introduced for explicitly utilizing the knowledge to advance the text response generation.
Extensive experiments on a public dataset verify the superiority of the proposed DKMD over state-of-the-art competitors.
\end{abstract}


% \ccsdesc[500]{Information systems~Multimedia information systems}


\maketitle

% \IEEEraisesectionheading{\section{Related Work}\label{sec:related-work}}
\section{Introduction}

\begin{figure}[!t]
    \centering
    \includegraphics[scale=0.32]{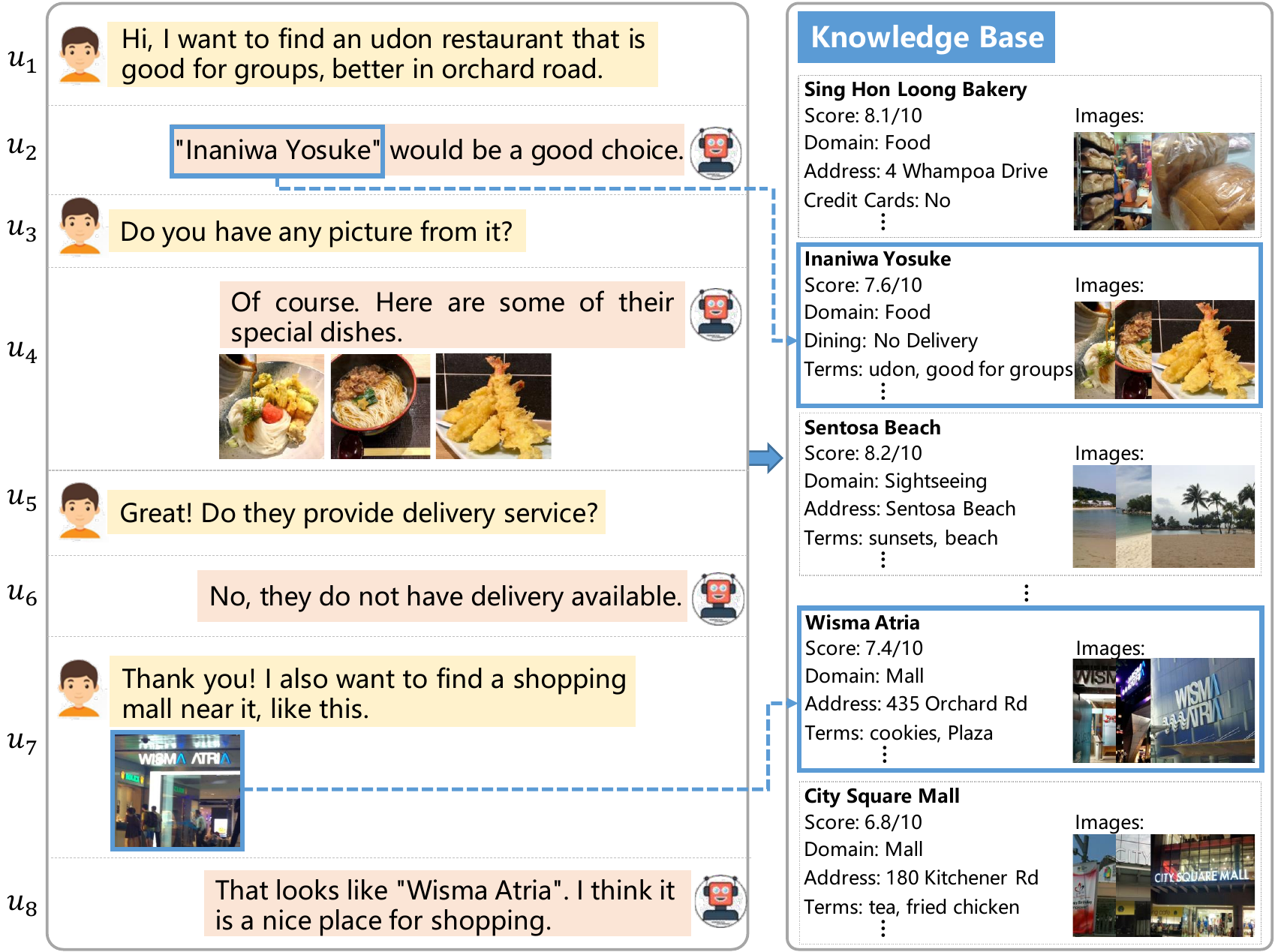}
    % \vspace{-0.5em}
    \caption{Illustration of a multimodal dialog system between a user and an agent. 
    ``u'': utterance.}
    \vspace{-1em}
    \label{example_MDialog}
\end{figure}

According to the report of Salesforce\footnote{https://startupbonsai.com/chatbot-statistics.}, roughly $68$\% of customers prefer dialog agents rather than waiting for human services because dialog agents can provide quick answers. {\color{black}Due to the substantial economic value, \mbox{task-oriented} dialog systems, which aim to conduct specific tasks in certain vertical domains, such as ticket booking and restaurant table reserving, have attracted increasing research attention.}
Although existing research efforts have attained impressive results, most of them work purely on the single-modality~(\emph{i.e.,} textual modality) dialog system, neglecting that both the user and the agent may need to employ certain visual clues~(\emph{i.e.,} images) to deliver their needs or services.
% As a matter of fact, visual cues not only enable the user intuitively express their intention but also
% allow the agent to better understand
% % contribute to the agent`s understanding of 
% the user`s intention~\cite{DBLP:conf/mm/ZhangLGLWN21}.
% cues which are of crucial importance to convey the user intention.
% In fact, visual cues not only 
As depicted in Figure~\ref{example_MDialog}, the agent shows special dishes for the user via images in the utterance  $u_4$, while the user describes his/her desired shopping mall with the image in the utterance $u_7$.
% As depicted in Figure~\ref{example_MDialog}, the user describes his/her desired food via the images in the utterance $u_3$, which stimulates the expression of the user`s intention and enables the agent to understand the user's desired food.
Therefore, multimodal \mbox{task-oriented} dialog systems merit our specific attention.

% In this work, we aim to investigate the multimodal task-oriented dialog systems, which have attracted many researchers` attention.
% In this work, we work towards the multimodal task-oriented dialog systems, which have attracted many researchers` attention.
In general, multimodal task-oriented dialog systems mainly involve two tasks~\cite{DBLP:conf/aaai/SahaKS18}: the text response generation and the image  response selection.
As compared with the image response selection task, the text response generation task is more challenging, whose performance is far from satisfactory. 
% In this work, 
% % similar to~\cite{DBLP:conf/mm/HeLLCXHY20}, 
% we particularly focus on the task of textual response generation 
% % conditioned on multimodal task-oriented dialog systems. 
% due to the fact that the image response selection task has been well addressed by previous studies, where the evaluation metric~(\emph{i.e.,} Recall@1) has achieved $99.83\%$~\cite{ma-etal-2022-unitranser}. 
Existing multimodal task-oriented dialog systems mainly adopt the \mbox{encoder-decoder} framework for text response generation. %, and devised various encoders to model the multimodal dialog context. 
In particular, recent studies have recognized the pivotal role of the knowledge base for multimodal dialog systems, and designed various schemes for incorporating knowledge to enhance the user's intention understanding~\cite{DBLP:conf/mm/ZhangLGLWN21,DBLP:journals/tip/NieJWWT21,DBLP:conf/acl/ChauhanFEB19,DBLP:conf/mm/LiaoM0HC18,DBLP:conf/sigir/CuiWSHXN19,DBLP:conf/mm/NieWHWT19,DBLP:conf/mm/HeLLCXHY20,ma-etal-2022-unitranser}. 
Although they have achieved significant progress,  these research efforts suffer from two key limitations. \mbox{1) \textbf{Overlook}} \textbf{the benefit of generative  pre-training.} Previous studies follow the conventional \mbox{train-from-scratch} paradigm and fail to leverage the generative  pre-training technique, ignoring the powerful text generation ability of generative pretrained language models~ (GPLMs)~\cite{DBLP:conf/naacl/DevlinCLT19,DBLP:journals/jmlr/RaffelSRLNMZLL20,DBLP:conf/acl/LewisLGGMLSZ20}.
% , which have achieved promising performance in various natural language generation tasks, such as the abstractive summarization~\cite{DBLP:conf/emnlp/YuDLF21} and the commonsense-aware text generation~\cite{DBLP:conf/emnlp/JiKHWZH20}
2) \textbf{Ignore the textual context related knowledge.}  % Previous studies have noticed the pivotal role played by the knowledge base of the dialog system for the user`s intention understanding and textual response generation~\cite{DBLP:conf/mm/ZhangLGLWN21,DBLP:journals/tip/NieJWWT21,DBLP:conf/acl/ChauhanFEB19,DBLP:conf/mm/LiaoM0HC18,DBLP:conf/sigir/CuiWSHXN19,DBLP:conf/mm/NieWHWT19,DBLP:conf/mm/HeLLCXHY20,ma-etal-2022-unitranser}. Nevertheless, %they all rely on the benchmark MMD dataset that only supports the agent to refer to the knowledge base according to the images provided by the user 
{\color{black}
Previous studies only select knowledge based on the visual context (\emph{e.g.,} the image associated with the utterance $u_7$ in Figure~\ref{example_MDialog}). Namely, they only involve the visual context related knowledge to enhance the user intention modeling. Nevertheless, they overlook that the textual context plays the dominant role in the dialog, and could also be used for fetching related knowledge from the knowledge base to enhance the text response generation.
}
To address these limitations, in this work, we target at comprehensively utilizing the multimodal context in knowledge selection with the backbone of GPLMs to improve the performance of text response generation for multimodal \mbox{task-oriented} dialog systems. 
This is, however, non-trivial due to the following three challenges. 
% As for the task of textual response generation, despite significant progress has been made  by existing efforts~\cite{DBLP:conf/aaai/SahaKS18,DBLP:conf/mm/NieWHWT19,DBLP:conf/mm/HeLLCXHY20,DBLP:conf/mm/ZhangLGLWN21}, they focus on the fashion domain and overlook the general domain whereby the dialog scenarios are more complex and practical. 
% To bridge the research gap, Liao et al.~\cite{DBLP:conf/sigir/LiaoLZHC21} build the multimodal  multi-domain conversational dataset, which pushes the frontiers of  general multimodal dialog systems. 
% To bridge the research gap, Liao et al.~\cite{DBLP:conf/sigir/LiaoLZHC21} built a new multimodal  conversational dataset spanning multiple domains.
% which pushes the frontiers of  multimodal dialog systems in general domain. 
% {\color{blue}Although the pioneer studies push the frontiers of  multimodal dialog systems in the general domain, they overlook the benefit of pre-training, which propels us to explore the potential of incorporating the GPLMs in the context of multimodal dialog systems.}
% Meanwhile, recent years have witnessed the tremendous strides of generative pretrained language models~(GPLMs), which propels us to explore the potential of incorporating the GPLMs in the context of multimodal dialog systems.
% \caption{Illustration of the proposed model, which consists of two vital components: dual knowledge-enhanced  context learning and knowledge-enhanced response generation.}
% In this work, we aim to investigate the task of textual response generation conditioned on task-oriented multimodal systems.
% where the GPLMs are fully explored.
% However, the textual response generation task of multimodal task-oriented dialog systems based on GPLMs is non-trivial due to the following challenges.
\mbox{1) The} multimodal dialog context cannot fit well with the GPLMs that are pretrained with only the textual corpus, and thus directly feeding the multimodal context into GPLMs deteriorates the text generation capability of GPLMs. Therefore, how to subtly adapt GPLMs to cope with the multimodal dialog context constitutes the main challenge.
% 2) In fact, there is complicated semantic relation hidden in multimodal dialog context.
% 2) In fact, the  knowledge is indispensable for the accurate textual response from both the user intention understanding and the explicit response generation.
% {\color{blue}{where different modality context may be both associated with knowledge.}}
% where it not only contributes to characterize the user intention but also can provide explicit clues for the textual generation. For example, as shown in Figure~\ref{example_MDialog}, the conversation between the user and the agent involve 
% For example, as shown in Figure~\ref{example_MDialog}, the textual context involves the semantic knowledge~(\emph{e.g.,} domain, location, and representative food), while the visual context may be relevant to knowledge entities which have similar food pictures. 
% In addition, the related knowledge of ``Wisma Atria'', especially the location, is instructive for the response~(\emph{i.e.,} $u_6$) generation.
% Apart from the role in capturing the user intention, the knowledge is also essential to explicitly promote the textual response generation.
% while the visual context may be relevant to images of entities in the knowledge base.
% while the visual context comes to the images of entities in the knowledge base. 
% As aforementioned, both the textual context and the visual context are  referred to select the multimodal context related knowledge.
\mbox{2) As} aforementioned,
% it is promising to refer  knowledge base from both the textual context and the visual context perspectives.
% Notably, 
the context related knowledge is of crucial importance to the text response generation. For example, as shown in Figure~\ref{example_MDialog}, the agent can generate the proper response~(\emph{e.g., }$u_6$) only conditioned on the attribute knowledge~(\emph{i.e., } ``No Delivery'') of ``Inaniwa Yosuke''.
Hence, how to accurately select the knowledge concerning the given multimodal context and properly inject knowledge to enhance the user intention modeling and text response generation with GPLMs
% facilitate the text response generation from both the intention understanding and explicit generation
is another crucial challenge. 
3) {\color{black}
Both textual context and visual context serve to demonstrate the user's intention, where they are closely related and mutually reinforce each other. As shown in Figure~\ref{example_MDialog}, the user demonstrates his/her intention of finding a restaurant and a shopping mall with not only the textual description (\emph{e.g.,} `an udon restaurant', `good for groups', `better in orchard road' and `shopping mall near it'), but also images of his/her desired shopping mall. To be specific, the user depicts that he/she tends to find a shopping mall via the textual context of the utterance $u_7$, and provides the corresponding image of the desired shopping mall by the visual context.}
% Both textual context and visual context serve to demonstrate the user's intention, where they are closely related and mutually reinforce each other. 
% % which can be accurately captured 
% % The textual context and the visual context complement each other, and both contribute to expressing the user's intention.
% As shown in Figure~\ref{example_MDialog},  the user demonstrates his/her intention of finding a restaurant and a shopping mall with not only the textual description~(\emph{e.g.,} `an udon restaurant', `good for groups', `better in orchard road' and `shopping mall near it'), but also images of his/her desired shopping mall.
% To be specific, the user depicts that he/she tends to find a shopping mall via the textual context of the utterance $u_7$, and provides the corresponding image of the desired shopping mall by the visual context.
% % describes his/her desired shopping mall with the image in the utterance 7.
% }
Therefore,  how to mine the context  cross-modality semantic relation based on GPLMs and thus accurately capture the user's intention is a tough challenge.

\begin{figure*}[!t]
    \centering
    \includegraphics[scale=0.54]{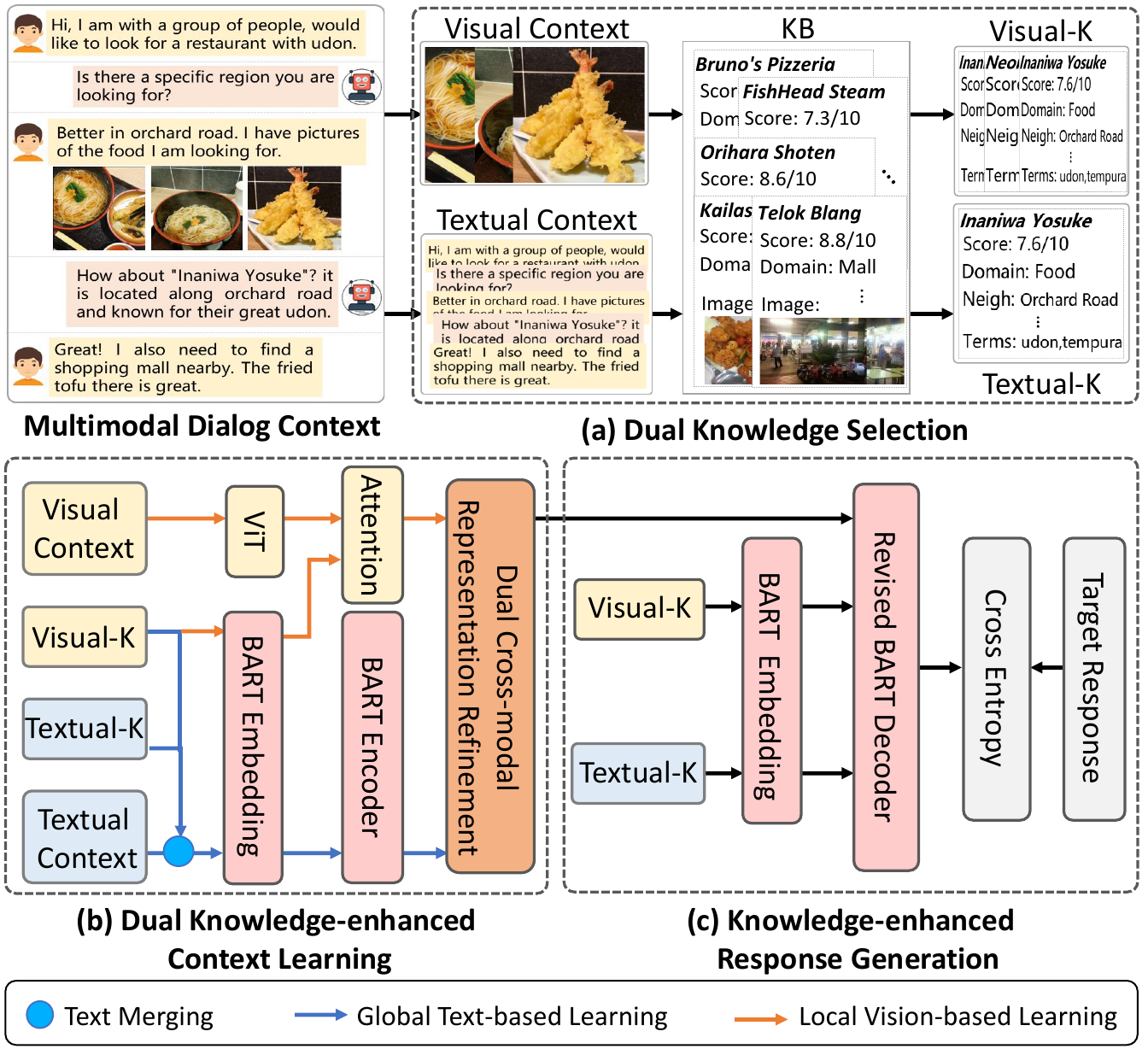}%Figure_Model_v7_14
    \vspace{-0.5em}
    \caption{Illustration of the proposed DKMD, which consists of three vital components: \emph{(a) Dual Knowledge Selection}, \emph{(b) Dual Knowledge-enhanced Context Learning}, and \emph{(c) Knowledge-enhanced Response Generation}. `KB': Knowledge base. `K': Knowledge.
    % ``T'', ``V'', ``KB'', and ``K'' refer to textual, visual, knowledge base and knowledge, respectively.
    % The bottom demonstrates the design details of three modules: local visual learning, dual refinement, and revised BART decoder.
    }
    \vspace{-0.5em}
    \label{Model}
\end{figure*}

To address the challenges mentioned above,  
we propose a novel dual
knowledge-enhanced generative pretrained language model for multimodal task-oriented dialog systems, DKMD for short,
% we propose a novel 
% dual knowledge-enhanced multimodal dialog system, DKMD for short,
% knowledge-enhanced dual generative pretrained model for task-oriented  multimodal dialog systems, DKMD for short,
% Knowledge-enhanced Dual Multimodal Dialog system, DKMD for short, 
where  BART~\cite{DBLP:conf/acl/LewisLGGMLSZ20} is adopted as the backbone.
% In particular, the proposed DKMD consists of two key components: \emph{dual knowledge-enhanced  context learning} and \emph{knowledge-enhanced response generation}.
As illustrated in Figure~\ref{Model}, DKMD contains three vital components: \emph{dual knowledge selection}, \emph{dual \mbox{knowledge-enhanced} context learning}, and \emph{knowledge-enhanced response generation}.
To be specific, the dual knowledge selection component is devised to select the context related knowledge from the whole knowledge base according to both the textual and visual modality of the given context.
% To be specific, the dual knowledge selection component is devised to select the dual multimodal context related knowledge from the whole knowledge base.
Thereafter, the dual \mbox{knowledge-enhanced} context learning component aims to properly incorporate dual knowledge (\textit{i.e.}, both textual and visual context related knowledge) to the multimodal context modeling and hence accurately captures the user's intention.
% To be specific, we first devise the dual knowledge selection module to select the multimodal context related knowledge from the whole knowledge base. 
In particular, considering different roles of multimodal context in conveying the user's intention, we design the \mbox{knowledge-enhanced} context representation module with the global \mbox{knowledge-enhanced} textual representation learning and local \mbox{knowledge-enhanced} visual representation
learning.
% knowledge-enhanced multimodal context learning
%that utilizes knowledge to enhance the multimodal context learning from both the global and local perspectives.
% Moreover, we introduce the dual cross-modal representation refinement module to capture the semantic relation hidden in the multimodal context and facilitate  the user intention modeling.
Moreover, we introduce the dual cross-modal representation refinement module, comprising
vision-oriented representation refinement and text-oriented
representation refinement, to capture the semantic relation hidden in the multimodal context and facilitate  the user intention modeling.
% , comprising \mbox{vision-oriented} representation refinement and text-oriented representation refinement.
% to capture the semantic relation hidden in the multimodal context and facilitate  the user intention modeling.}
% Thereafter, we conduct the Knowledge-enhanced Context Representation, where the knowledge is utilized to enhance the context learning and the dual semantic relation hidden in the multimodal context is also fully modeled. 
% In addition, to explicitly use the knowledge to advance the text response generation, 
Ultimately, 
the \mbox{knowledge-enhanced} response generation component targets at explicitly using the knowledge to advance the text response generation, where a revised  BART decoder with an additional  \mbox{dot-product} knowledge-decoder attention~(DKDA) \mbox{sub-layer} is introduced.
Extensive experiments on one public dataset have fully validated the effectiveness of our proposed DKMD.

Our main contributions can be summarized as follows:
\begin{itemize}
    \item To the best of our knowledge, we are among the first to integrate the GPLMs into multimodal \mbox{task-oriented} dialog systems.
    In particular, we propose a novel dual \mbox{knowledge-enhanced} generative pretrained language model for the text response generation task.
    % , where we propose a novel dual knowledge-enhanced generative pretrained language model 
    % In particular, we propose a novel  conduct the dual knowledge selection by referring to the knowledge base from both the textual and visual contexts perspectives.
    % In particular, we conduct the dual knowledge selection by referring to the knowledge base from both the textual and visual contexts perspectives.
    % In particular, we conduct the dual knowledge selection by referring to the knowledge base from both the textual context and the visual context perspectives.
    
    % where the dual knowledge is selected  and properly injected to enhance the user intention modeling and textual response generation.}
    % To the best of our knowledge, we are among the first to incorporate the GLPMs into multimodal task-oriented dialog systems where the  semantic relation hidden in the multimodal context is fully explored.
    \item We propose the dual knowledge-enhanced context learning component, which seamlessly integrates the selected dual knowledge into the multimodal context learning from global and local perspectives and also explores the context cross-modality semantic relation to facilitate the user intention modeling.   
    % We propose the knowledge-enhanced context representation module,
    % which seamlessly integrates the knowledge  from global and local perspectives, and devise the knowledge-enhanced decoder that can utilize knowledge to stimulate the precise textual response generation explicitly. Besides, we  explore the semantic relation hidden in the multimodal context to further facilitate the user intention modeling.   
    \item  We devise the knowledge-enhanced decoder that can utilize knowledge to stimulate the precise text response generation explicitly. 
    Extensive experiments on the public dataset demonstrate the effectiveness of our proposed DKMD.
    As a byproduct, we have released codes and involved parameters to facilitate the research community\footnote{https://multimodaldialog.wixsite.com/website.}.    
\end{itemize}

% \IEEEraisesectionheading{\section{Related Work}\label{sec:related-work}}
\section{Related Work}

% \IEEEPARstart{O}{ur} work is related to the studies of multimodal dialog systems and pretrained language models.
% \IEEEPARstart{O}{ur} work is related to the studies of multimodal dialog systems and pretrained language models.

% In this section, we briefly introduce the studies of unimodal dialog systems, multimodal dialog systems and pretrained language models, which are closely related to our work.
In this section, we briefly introduce the studies of \mbox{task-oriented} dialog systems and pretrained language models, respectively.

\subsection{Task-oriented Dialog Systems}
{\color{black}Over the last few years, task-oriented dialog systems, which aim  to conduct specific tasks in a certain domain, such as restaurant table reserving and sightseeing ticket booking, have penetrated into many aspects of our lives and attracted a surge of research.}

% Task-oriented dialog systems aim to conduct specific tasks in a certain domain, such as restaurant table reserving and sightseeing ticket booking.
Traditional task-oriented dialog systems mainly adopt a pipeline structure and usually contain the following functional modules: natural language understanding, dialogue state tracking, policy learning, and natural language generation.
Specifically, the natural language understanding module aims to classify the user's intentions, and then the dialogue state tracking module can track the current state and fill in the predefined slots.
Thereafter, the policy learning module predicts the following action on the basis of the current state representation, and the natural language generation module returns the response through generation methods~\cite{DBLP:conf/sigir/LiLW21,10.5555/2969033.2969173,DBLP:journals/tkde/LiaoTMYHC22,DBLP:journals/tkde/LiZLC22,DBLP:journals/tois/DengLZDL22} or predefined templates.
% DBLP:conf/nips/ShiCWYWW15,DBLP:journals/corr/ChungGCB14
Despite the remarkable success of the pipeline-based methods, they are prone to suffer from error propagation~\cite{DBLP:conf/acl/KanHLJRY18} and heavy dependence on the sequential modules~\cite{DBLP:conf/aaai/ZhangOY20}. 
% In particular, some dialog systems are dedicated to absorbing domain knowledge to achieve notable improvement~\cite{DBLP:conf/emnlp/ChenLZDCYT19,DBLP:conf/eacl/Rojas-BarahonaG17}.

With the evolution of deep neural networks, several efforts have been made toward  \mbox{end-to-end} task-oriented dialog systems~\cite{DBLP:conf/ijcnlp/LiCLGC17,DBLP:conf/coling/WangLBLHXY20,9699426,DBLP:journals/tois/RenCRKMR21,DBLP:conf/aaai/LiYQ22,DBLP:conf/aaai/YoungCCZBH18,DBLP:conf/acl/FungWM18}. 
{\color{black}
Li et al.~\cite{DBLP:conf/aaai/LiYQ22}  studied the emotion recognition task conditioned on text-based dialog systems, where BART is adopted as the backbone. In particular, this method designs the auxiliary response generation task to further enhance the context understanding ability. Young et al.~\cite{DBLP:conf/aaai/YoungCCZBH18} worked on retrieval-based text-based dialog systems, and utilized the LSTM network to encode the commonsense knowledge. In addition, Madotto et al.~\cite{DBLP:conf/acl/FungWM18} proposed the memory-to-sequence model for the textual response generation task, where the multi-hop attention mechanism and pointer networks are incorporated. Although these efforts have achieved compelling success, they focus on the pure textual modality, i.e., the \mbox{single-modality} task-oriented dialog system, neglecting the visual modality. In reality, both the user and the agent may need to refer to certain images to deliver their needs or services. 
% Li et al.~\cite{DBLP:conf/aaai/LiYQ22} studied the emotion recognition task conditioned on the text-based dialog system, where BART is adopted as the backbone. In particular, this method integrates the auxiliary response generation task to further enhance the model’s ability of context understanding.
% Young et al.~\cite{DBLP:conf/aaai/YoungCCZBH18} worked on retrieval-based text-based dialog systems, and utilized LSTM network to encoder the commonsense knowledge.
% In addition, Madotto et al.~\cite{DBLP:conf/acl/FungWM18} proposed the  memory-to-sequence model for the textual response generation task, where the multi-hop attention mechanism and pointer networks are incorporated.
% Although these efforts have achieved compelling success, they focus on the pure textual modality, \emph{i.e.,} the single-modality task-oriented dialog system, neglecting the visual modality. 
% }
% However, the above research efforts on task-oriented dialog systems focus on the textual modality neglecting visual cues, which cannot comprehensively characterize the user's intention and is still far from satisfactory.
% In reality, both the user and the agent may need to refer to certain images to deliver their needs or services. 
In particular, Saeid et al.~\cite{DBLP:conf/wsdm/KordanK18} investigated the synergistic effect of text and image in the multimodal retrieval task, which verifies that the text and image tend to be complementary and can be considered as one unit.
}
% In particular, Saeidet al.~\cite{DBLP:conf/wsdm/KordanK18} investigated that there is the synergistic effect of text and images, which further verifies the necessity of working on  multimodal dialog systems.}
% which is deserved to be discussed. }
Therefore,
Saha et al.~\cite{DBLP:conf/aaai/SahaKS18} investigated the multimodal dialog system, and proposed a multimodal hierarchical \mbox{encoder-decoder} model~(MHRED) for addressing the two primary tasks of the multimodal dialog system: text response generation and image
response selection. 
Moreover, they released a large-scale multimodal dialog dataset in the context of online fashion shopping, named MMD, which significantly promotes the research progress on multimodal dialog systems.
% Saha et al.~\cite{DBLP:conf/aaai/SahaKS18} built a large-scale multimodal dialog dataset~(\emph{i.e.,} MMD) {\color{blue}{in the context of online fashion shopping.}}
% which drives the progress of multimodal dialog systems. 
% They formulated the multimodal dialog systems with two basic tasks: textual response generation and image response selection, and proposed a multimodal hierarchical encoder-decoder model (MHRED) for addressing them. 
% they also proposed a multimodal hierarchical encoder-decoder model~(MHRED) and  defined two basic tasks~(\emph{i.e.,} textual response generation and image response selection).
In particular, several efforts further  explore the semantic relation in the multimodal dialog context and incorporate  knowledge based on the framework of  MHRED~\cite{DBLP:conf/mm/ZhangLGLWN21,DBLP:journals/tip/NieJWWT21,DBLP:conf/acl/ChauhanFEB19,DBLP:conf/mm/LiaoM0HC18,DBLP:conf/sigir/CuiWSHXN19,DBLP:conf/mm/NieWHWT19}.
% Thereafter, several efforts refines MHRED by exploring the semantic relation in the multimodal dialog context and incorporating domain knowledge~\cite{DBLP:conf/mm/ZhangLGLWN21,DBLP:journals/tip/NieJWWT21,DBLP:conf/acl/ChauhanFEB19,DBLP:conf/mm/LiaoM0HC18,DBLP:conf/sigir/CuiWSHXN19,DBLP:conf/mm/NieWHWT19}.
For example, Liao et al.~\cite{DBLP:conf/mm/LiaoM0HC18} developed a taxonomy-based visual semantic learning module to capture the fine-grained semantics~(\emph{e.g.,} the category and attributes of a product) in product images, and introduced a memory network to integrate the knowledge of fashion style tips.
In addition, Nie et al.~\cite{DBLP:conf/mm/NieWHWT19} devised a multimodal dialog system with multiple decoders, which can  generate diverse responses according to the user's intention and adaptively integrate the related knowledge.
Recently, 
% with the flourish of transformer~\cite{DBLP:conf/nips/VaswaniSPUJGKP17},
some studies have resorted  to Transformer~\cite{DBLP:conf/nips/VaswaniSPUJGKP17} to investigate the multimodal dialog systems~\cite{DBLP:conf/mm/HeLLCXHY20,ma-etal-2022-unitranser} due to its impressive results in natural language processing~(NLP)  tasks~\cite{DBLP:conf/naacl/DevlinCLT19,DBLP:journals/jmlr/RaffelSRLNMZLL20,DBLP:conf/acl/LewisLGGMLSZ20,DBLP:journals/tkde/ChenZLYJ22,9226110}.
For example, He et al.~\cite{DBLP:conf/mm/HeLLCXHY20} introduced a Transformer-based element-level encoder, which can capture the semantic dependencies of multimodal elements~(\emph{i.e.,} words and images) via the attention mechanism.
% , and also determinate 
% leverage related images from the whole dialogue history to 
% and leverage related images from 
% relevancy between each image in the context and the current turn.

% Currently, the image response selection task of multimodal dialog systems has been well addressed, where the related evaluation metric~(\emph{i.e.,} Recall@1) has achieved $99.83\%$~\cite{ma-etal-2022-unitranser}, while the textual response generation task still remains challenging.
As compared with the image response selection task, the  text response generation task is more challenging, whose performance is far from satisfactory. 
Therefore, in this work, we particularly study the task of text response generation in the context of multimodal \mbox{task-oriented} dialog systems.
% the textual response generation task in the context of multimodal task-oriented dialog systems. 
Notably, although the pioneer studies have achieved tremendous strides on this task, they overlook the benefit of generative pre-training and only utilize the attribute knowledge concerning the visual context of the dialog.
Beyond that, in this work, we aim to generate a precise response  
% based on the  dual knowledge selection 
by utilizing  pretrained techniques and capturing related knowledge from both the textual context and visual context perspectives.

\subsection{Pretrained Language Models}
As an emerging technique, pretrained language models have been arresting much research attention~\cite{DBLP:conf/emnlp/YuDLF21,DBLP:conf/naacl/DevlinCLT19,DBLP:journals/jmlr/RaffelSRLNMZLL20,DBLP:conf/acl/LewisLGGMLSZ20,9372844,DBLP:journals/tois/MustarLP22} and achieve remarkable success in plenty of NLP tasks.
Initially, Word2vec~\cite{10.5555/2999792.2999959} and GloVe~\cite{pennington-etal-2014-glove} are proposed to obtain pretrained word embeddings based on shallow architectures.
Thereafter, with the flourish of Transformer, considerable studies make efforts to devise \mbox{Transformer-based} pretrained models~\cite{DBLP:conf/naacl/DevlinCLT19,DBLP:journals/jmlr/RaffelSRLNMZLL20,DBLP:conf/acl/LewisLGGMLSZ20}.
For example, Devlin et al.~\cite{DBLP:conf/naacl/DevlinCLT19} proposed the bidirectional encoder representation from transformer (BERT) to capture the accurate textual representation via two pre-training tasks: masked language model and next sentence prediction.
Since then, BERT has achieved compelling success in many tasks
% ~\cite{DBLP:journals/jmlr/RaffelSRLNMZLL20,DBLP:conf/sigir/ChatterjeeD22,DBLP:conf/sigir/ChenSPFN21,DBLP:conf/sigir/XuCCW20,DBLP:conf/sigir/CaiZNL20,DBLP:journals/tois/FrummetEL22}
, such as entity ranking~\cite{DBLP:conf/sigir/ChatterjeeD22,DBLP:conf/sigir/ChenSPFN21}, semantic reasoning~\cite{DBLP:conf/sigir/XuCCW20,DBLP:journals/tois/FrummetEL22}, and machine reading comprehension ~\cite{DBLP:conf/sigir/CaiZNL20,DBLP:journals/jmlr/RaffelSRLNMZLL20}.
For example, Chen et al.~\cite{DBLP:conf/sigir/ChenSPFN21} utilized BERT to extract the textual representation from users' heterogeneous \mbox{multi-modal} posts, to promote the user representation learning and hence promote user identity linkage across multiple social networks.
In addition, Chatterjee et al.~\cite{DBLP:conf/sigir/ChatterjeeD22}  devised the query-specific BERT entity representations by fine-tuning BERT, which is instructive for the entity ranking task.
% Furthermore, 
What is more, Lewis et al.~\cite{DBLP:conf/acl/LewisLGGMLSZ20} presented a \mbox{Transformer-based} denoising autoencoder~(BART) for the language generation task, with the bidirectional encoder and the autoregressive decoder.
{\color{black} 
Similarly, Radford et al.~\cite{Radford2018ImprovingLU} proposed a semi-supervised approach for language understanding tasks with a combination of unsupervised pre-training and supervised fine-tuning. Raffel et al.~\cite{DBLP:journals/jmlr/RaffelSRLNMZLL20} incorporated transfer learning techniques and proposed the text-to-text transfer transformer (T5) in the natural language processing field.
% Similarly, Radford et al.~\cite{Radford2018ImprovingLU} explored a semi-supervised approach for language understanding tasks with a combination of unsupervised pre-training and supervised fine-tuning.
% Raffel et al.~\cite{DBLP:journals/jmlr/RaffelSRLNMZLL20} incorporated  transfer learning techniques and proposed the text-to-text transfer transformer~(T5) in the natural language
% processing field.
}
% to derive the progress of the  generation
With the remarkable progress of generative pretrained language models, a surge of \mbox{follow-up} works~\cite{DBLP:conf/emnlp/YuDLF21,DBLP:conf/sigir/SongJLZCN22,DBLP:conf/sigir/KangLCJ21,DBLP:conf/sigir/ZhuYGZ021,DBLP:journals/tois/BahrainianZCE22} solve diverse tasks by adapting publicly available pretrained language models.
For example, Yu et al.~\cite{DBLP:conf/emnlp/YuDLF21} designed a vision-guided generative pretrained language model based on BART and T5 for the multimodal abstractive summarization task.
Likewise, Song et al.~\cite{DBLP:conf/sigir/SongJLZCN22} proposed a vision-to-prompt based multi-modal product summary generation framework, where BART is adopted as the backbone.
{\color{black}
% Kang et al.~\cite{DBLP:conf/sigir/KangLCJ21}  proposed a sequential recommendation model on the basis of BART, which entangles bidirectional encoder and \mbox{auto-regressive} decoder with noisy transformations for user interaction.
Kang et al.~\cite{DBLP:conf/sigir/KangLCJ21} proposed a sequential recommendation model on the basis of BART, which entangles bidirectional encoder and auto-regressive decoder with noisy transformations for user interaction.
}
% ~\cite{DBLP:conf/emnlp/YuDLF21} multimodal abstractive summarization task

Although generative pretrained language models have shown compelling success in many tasks, limited efforts have been devoted to conducting the text response generation in multimodal \mbox{task-oriented} dialog systems.
% Toward this end, 
To fill the research gap, we adapt the publicly available pretrained BART  to integrate the multimodal context and corresponding knowledge to enhance the response generation capability of our model.
% are seamlessly integrated.

% which can exert the rich knowledge from large-scale corpus to fulfill the text response generation task conditioned on multimodal dialog systems.

% \IEEEraisesectionheading{\section{Related Work}\label{sec:related-work}}

% \begin{table}[!t]
%     \centering
%   %  \vspace{-1em}
%     \caption{Detailed statistics of the MMConv dataset.}
%     % \setlength{\tabcolsep}{7mm}{
%     \begin{tabular}{|l||r|}
%     \hline
%     Entry                                  & Number         \\ \hline
%     \#dialogues                            & $5,106$          \\ \hline
%     \#turns                               & $39,759$       \\ \hline
%     \#single modality dialogues             & $751$  \\ \hline
%     \#multi-modality dialogues              & $4,355$ \\ \hline  
%     \#single domain dialogues             & $808$  \\ \hline
%     \#multi-domain dialogues              & $4,298$ \\ \hline       
%     \#total entities in the knowledge base  & $1,771$         \\ \hline
%     % \#total images                         & $113,953$        \\ \hline
%     \end{tabular}
%     \vspace{-2em}
%     \label{dataset}
% \end{table}

% \begin{comment}
\begin{table}[!t]
  \centering
  \caption{Summary of the main notations.}
  \vspace{-1em}
  \setlength{\tabcolsep}{5mm}{
  \begin{tabular}{|c||l|}
      \hline
      Notation & Explanation \\
      % \hline
      \hline
      $\mathcal{D}$ & The set of training dialog pairs. \\
      \hline
      $\mathcal{C}_i$ & The multimodal context of the $i$-th sample  in the training dialog $\mathcal{D}$. \\
      \hline
      $\mathcal{R}_i$ & The target textual response of the $i$-th sample in the training dialog $\mathcal{D}$. \\
      \hline
      $\mathcal{T}_i$ & The set of  sequence of historical textual utterances in $\mathcal{C}_i$. \\
      \hline 
      $t_g^i$ & The $g$-th token in $\mathcal{T}_i$. \\
      \hline          
      $\mathcal{V}_i$ & The set of images in $\mathcal{C}_i$. \\
      \hline  
      $v_j^i$ & The $j$-th image in $\mathcal{V}_i$. \\
      \hline
      $r_n^i$ & The $n$-th token in $\mathcal{R}_i$. \\
      \hline        
      $\mathcal{K}$ & The knowledge base. \\
      \hline    
      $e_p$ & The $p$-th entity in the knowledge base $\mathcal{K}$. \\
      \hline       
      $\mathcal{A}_p$ & The attribute knowledge of $e_p$ in the knowledge base $\mathcal{K}$. \\
      \hline   
      $\mathcal{I}_p$ & The images of $e_p$ in the knowledge base $\mathcal{K}$. \\
      \hline       
      $\mathbf{T}_E$ & The final learned knowledge-enhanced context representation. \\
      \hline     
      $\tilde{\mathbf{y}}$ & The predicted token distribution. \\
      \hline         
  \end{tabular}}
  \vspace{-0.5em}
  \label{tab:notation}
\end{table}
% \end{comment}

\section{Preliminary}
Formally, we first declare some notations. In particular, we use bold uppercase letters (\emph{e.g.,} $\mathbf{X}$) and bold lowercase letters (\emph{e.g.,} $\mathbf{x}$) to represent matrices and vectors, respectively. We employ \mbox{non-bold} letters (\emph{e.g.,} $x$) to denote scalars and Greek letters (\emph{e.g.,} $\gamma$) to parameters. If not clarified, all vectors are in column forms. 
The main notations used in this article are summarized in Table~$\ref{tab:notation}$.
% 加字母表
% \subsection{BART}

We choose BART as our backbone for the text response generation due to its superior performance in many text generation tasks, such as  abstractive summarization~\cite{DBLP:conf/emnlp/YuDLF21} and community question answering~\cite{DBLP:conf/wsdm/GaoZWDC0022}.
In particular, BART is a \mbox{Transformer-based} denoising autoencoder,
% built with a sequence-to-sequence model, 
consisting of a position-wise embedding layer, a bidirectional encoder, and an autoregressive decoder.

\textbf{Position-wise Embedding Layer.}  
% For a given text $t$, the position-wise embedding layer first converts it into a sequence of tokens $[x_1,x_2, \cdots, x_M]$, where $M$ is the total number of tokens in the text.
Suppose we have a text  $t = [x_1,x_2, \cdots, x_M]$, where $x_q$ represents the $q$-th token and $M$ is the total number of tokens in the text.
Each token $x_q$ is assigned with an initial embedding $\mathbf{e}_q$ by a linear transformation as follows,

% The input text is first tokenized and converted to a sequence of tokens. Furthermore, each token $x_q$ can gain its initial embedding $\mathbf{e}_q$ with a linear transformation as follows,
\begin{equation}
    \mathbf{e}_q = \mathbf{W}^\top_{e}\mathbf{g}_q, 
    q=1,2,\cdots,M,
    % q\in\{1,2,\cdots,M\},
     \label{eq1_pre}
 \end{equation}
% where $\mathbf{W}_{\mathcal{U}} \in  {\mathbb{R}^{{|\mathcal{U}|}\times{D}}}$ is the token embedding matrix to be fine-tuned, $|\mathcal{U}|$ is the number of tokens in the token vocabulary and $D$ is the dimension of the token embeddings.
where $\mathbf{W}_{e} \in  {\mathbb{R}^{{|\mathcal{U}|}\times{D}}}$ is the token embedding matrix to be fine-tuned, $|\mathcal{U}|$ is the number of tokens in the token vocabulary $\mathcal{U}$, and $D$ is the dimension of the token embeddings.
$\mathbf{g}_q \in \mathbb{R}^{|\mathcal{U}|}$ is the one-hot vector, indicating the index of $x_q$ in the token vocabulary. 

To encode the order information among input tokens, position encodings~\cite{DBLP:journals/corr/VaswaniSPUJGKP17} are further inserted as follows,
\begin{equation}
    \mathbf{Z}_{0}^{enc}=[\mathbf{e}_1; \mathbf{e}_2; \cdots; \mathbf{e}_{M}]^\top + \mathbf{E}_{pos}, 
     \label{eq2_pre}
 \end{equation}
where $\mathbf{E}_{pos} \in \mathbb{R}^{{M}\times{D}}$ is the positional embedding matrix, each row of which corresponds to a token in the given text.  $\mathbf{Z}_{0}^{enc}\in \mathbb{R}^{{M}\times{D}}$ is the matrix containing all the final embeddings of tokens in the input text. $[;]$ refers to the concatenation operation.

% \textbf{Bidirectional Encoder.} The bidirectional encoder $\mathcal{B}_e$ is composed of $L$ encoder layers, which aims to encode the given text into the corresponding representation.
\textbf{Bidirectional Encoder.} The bidirectional encoder of BART, denoted as  $\mathcal{B}_e$, is composed of $L$ encoder layers, and used to 
encode the input text.
% encode the given text into the corresponding representation.
To be specific, each layer has two sub-layers: \mbox{1) multi-head} self-attention mechanism~(MSA), which aims to model the semantic dependencies among tokens in the input text; and \mbox{2) feed-forward} network~(FFN), used for the nonlinear transformation. 
Notably, each \mbox{sub-layer} is  followed by a residual connection and layer normalization~(LN) operations to enhance the model generalization  as follows,
%  \begin{equation}
%     \begin{split}
%     \begin{cases}
%     \mathbf{Z}_l^{S} = LN(MSA(\mathbf{Z}_{l-1}^{enc})+\mathbf{Z}_{l-1}^{enc}),\\
%      & l\in\{1,2,\cdots,L\},\\
%     \mathbf{Z}_l^{enc} = LN(FFN(\mathbf{Z}_l^{S})+\mathbf{Z}_l^{S}), 
%     \end{cases}
%     \end{split}
%     \label{eq3_pre}
%  \end{equation}

% \begin{flalign*}
\begin{equation}
% \begin{align*}
    \left\{
        \begin{array}{l}
            \mathbf{Z}_l^{S} =LN(MSA(\mathbf{Z}_{l-1}^{enc})+\mathbf{Z}_{l-1}^{enc}), \\  
            \mathbf{Z}_l^{enc} =LN(FFN(\mathbf{Z}_l^{S})+\mathbf{Z}_l^{S}),
        \end{array}
    \mbox{$l=1,2,\cdots,L,$}
    \right.
    % \end{align*}
    \label{eq3_pre}
\end{equation}
% \end{flalign*}
 where  $\mathbf{Z}_{l}^{enc}\in \mathbb{R}^{{M}\times{D}}$ refers to the output of  $l$-th encoder layer, and $\mathbf{Z}_{0}^{enc}$ is obtained by the aforementioned \mbox{position-wise} embedding layer in Eqn.~(\ref{eq2_pre}). $\mathbf{Z}_l^{S}\in \mathbb{R}^{{M}\times{D}}$ is the intermediate output of MSA in the $l$-th encoder layer. 
%  Ultimately, we treat the output of the $L$-th layer as the final encoded context representation, namely $\mathbf{Z}_L^{enc}\in \mathbb{R}^{{M}\times{D}}$. 
 Ultimately, the output of the $L$-th layer is treated as the final encoded context representation, namely $\mathbf{Z}_L^{enc}\in \mathbb{R}^{{M}\times{D}}$. 
% $l=\{1,2,\cdots,L\}$,

\textbf{Autoregressive Decoder.} The decoder $\mathcal{B}_d$ of BART also contains  $L$ decoder layers, which can generate the response based on the encoded representation.
To be specific, each layer  consists of three sub-layers: \mbox{1) masked} multi-head \mbox{self-attention} mechanism~(MMSA),
combined the mask mechanism and the operation making the output embeddings offset by one position, which ensures that the current output only depends on the known outputs;
% utilizing the mask mechanism to prevent positions from attending to subsequent positions. In addition, the output embeddings are offset by one position, which thus ensures that the current output only depends on the known outputs;
% which utilizes the mask mechanism to prevent positions from attending to subsequent positions and thus ensures that the current output only depends on the known outputs;
% previous outputs; 
\mbox{2) multi-head} encoder-decoder attention mechanism~(MEDA), which can distinguish the informative output of the encoder and adaptively assign weights to different previous outputs; 
% 2) Multi-head Encoder-Decoder Attention mechanism~(MEDA), which aims to perform multi-head attention over the output of the encoder; 
 and 3) FFN. Similar to the encoder, each sub-layer is  followed by a residual connection and layer normalization operations as follows,
 \begin{equation}
    \begin{split}
    \begin{cases}
    % \begin{align*}
    \mathbf{q}_l^{S} = LN(MMSA(\mathbf{q}_{l-1}^{dec})+\mathbf{q}_{l-1}^{dec}),\\
    \mathbf{q}_l^{E} = LN(MEDA(\mathbf{q}_l^{S}, \mathbf{Z}_L^{enc})+\mathbf{q}_l^{S}), & l=1,2,\cdots,L,\\
    \mathbf{q}_l^{dec} = LN(FFN(\mathbf{q}_l^{E})+\mathbf{q}_l^{E})
    % \end{align*}
    \end{cases}
    \end{split}
    \label{eq4_pre}
 \end{equation}
where $\mathbf{q}_l^{S}\in \mathbb{R}^{{D}}$ and $\mathbf{q}_l^{E}\in \mathbb{R}^{{D}}$ refer to the intermediate output of MMSA and MEDA in the $l$-th decoder layer, respectively.
$\mathbf{q}_{l}^{dec}\in \mathbb{R}^{{D}}$ denotes the final output of $l$-th decoder layer.  
Thereafter, the decoder $\mathcal{B}_d$ of BART employs the linear transformation and softmax function to project the decoder output into the  probability  space as follows,
 \begin{equation}
    \tilde{\mathbf{y}} = softmax(\mathbf{q}_{L}^{dec}\mathbf{W}_{y}+\mathbf{b}_y),
    \label{pre_predict}
 \end{equation}
where $\mathbf{W}_{y}$ and $\mathbf{b}_y$ represent the weight matrix  and bias vector, respectively.
$\tilde{\mathbf{y}}\in \mathbb{R}^{|\mathcal{U}|}$ denotes the predicted token distribution.
% Simultaneously, 
The predicted token of the current time step can be obtained according to the largest element of $\tilde{\mathbf{y}}$.

% $\tilde{\mathbf{y}}_n\in \mathbb{R}^{|\mathcal{U}|}$ for the $n$-th token, and thus obtains the $n$-th token $\tilde{r}_n$ according to the largest element of $\tilde{\mathbf{y}}_n$.

% $l=\{1,2,\cdots,L\}$, 
% In particular, MMSA can prevent positions from attending to subsequent positions and thus ensure the decoder work in an auto-regressive manner. Besides, MEDA can promote the encoded context incorporation via using the decoder embedding to attend over the context representation.

% \IEEEraisesectionheading{\section{Model}\label{sec:model}}
\section{Model}
In this section, we first formulate the research task of text response generation in multimodal \mbox{task-oriented} dialog systems, and then detail the proposed model illustrated in Figure~\ref{Model}, which comprises three vital components: \emph{dual knowledge selection}, \emph{dual knowledge-enhanced  context learning}, and \emph{\mbox{knowledge-enhanced} response generation}.

\subsection{Problem Formulation}

{\color{black}
As aforementioned above, in this work, we focus on the text response generation task conditioned on multimodal task-oriented dialog systems.}
% As compared with the image response selection task, 
% Considering the fact that
% As aforementioned above, compared with the image response selection task, 
% the text response generation task is more challenging, whose performance is far from satisfactory. 
% Therefore, in this work, we aim to investigate the  text response generation task conditioned on multimodal task-oriented dialog systems.}
Suppose  we have a set of $N$ training dialog pairs $\mathcal{D}=\{(\mathcal{C}_1, \mathcal{R}_1), (\mathcal{C}_2, \mathcal{R}_2), \cdots, (\mathcal{C}_{N}, \mathcal{R}_{N})\}$, where each pair comprises a multimodal  dialog context $\mathcal{C}_i$~(\emph{i.e.,} sequence of historical dialog utterances between the user and the agent) and a target text response $\mathcal{R}_i$.
% $N$ indicates the number of dialog pairs.
Notably, apart from the common textual modality, each utterance in $\mathcal{C}_i$ can also involve certain related images, as the user/agent may sometimes use images to facilitate the request/response expression. 
Accordingly, each multimodal  dialog context $\mathcal{C}_i$ can be decomposed into  a sequence of historical textual utterances $\mathcal{T}_i =[t_g^i]^{N_T^i}_{g=1}$~(\emph{i.e.,} a sequence of tokens) and a set of images $\mathcal{V}_i=\{v_j^i\}^{N_V^i}_{j=1}$, 
% Accordingly, each multimodal  dialog context $\mathcal{C}_i$ can be grouped into two subsets according to their modalities: the set of historical textual utterances $\mathcal{T}_i =\{t_g^i\}^{N_T^i}_{g=1}$~(\emph{i.e.,} a sequence of tokens) and the set of images $\mathcal{V}_i=\{v_j^i\}^{N_V^i}_{j=1}$, 
% where  refers to the historical textual utterances, \emph{i.e.,} a sequence of tokens and  stands for the set of images in the context, respectively.
% To be specific, suppose that we have a multimodal dialog context, \emph{i.e.,} sequence of historical dialog utterances between the user and the agent. Notably, apart from the common textual modality, each utterance can also involve certain related images, as the user/agent may sometimes use images to facilitate the request/response expression. 
% Formally, let $\mathcal{C}=\{\mathcal{T}, \mathcal{V}\}$ denote the given multimodal dialog context, where $\mathcal{T}=\{t_i\}^{N_T}_{i=1}$ refers to the historical textual utterances, \emph{i.e.,} a sequence of tokens and $\mathcal{V}=\{v_j\}^{N_V}_{j=1}$ stands for the set of images in the context, respectively.
where $t_g^i$ is the $g$-th token and $v_j^i$ is the $j$-th image of $\mathcal{C}_i$.  $N_T^i$ and $N_V^i$ refer to the total number of tokens and images, respectively.
Notably,  $N_V^i$ may be zero, \emph{i.e.,} there is no image in the  context $\mathcal{C}_i$.
The target text response of $\mathcal{C}_i$ can be denoted as $\mathcal{R}_i=[r_n^i]^{N_R^i}_{n=1}$, where $r_n^i$ denotes the $n$-th token and $N_R^i$ is the total number of tokens in the response.

Besides, the multimodal dialog system is equipped with a knowledge base, containing rich knowledge of $N_K$  entities $\mathcal{K} = \{e_p\}^{N_K}_{p=1}$. 
Specifically, for each entity $e_p$, the knowledge base provides a set of attributes $\mathcal{A}_p$ and images $\mathcal{I}_p$ characterizing it.
% each entity $e_p$ in the knowledge base is denoted as a key-value pair where the key refers to the entity name and the value stands for its corresponding knowledge~(\emph{i.e.,} a set of attributes $\mathcal{A}_p$ and a set of images $\mathcal{I}_p$).
% Specifically, each entity $e_p$ in the knowledge base is denoted as a key-value pair where the key refers to the entity name and the value stands for its corresponding knowledge~(\emph{i.e.,} a set of attributes $\mathcal{A}_p$ and a set of images $\mathcal{I}_p$).
% Specifically, each entity $e_p$ in the knowledge base is associated with a set of attributes $\mathcal{A}_p$ and a set of images $\mathcal{I}_p$.
The attributes~(\emph{e.g.,} score, domain, and location) reveal the semantic information of the entity, while the images  intuitively describe the entity, like the photos showing the appearance or food of a restaurant entity.
In a sense, we aim to devise a novel model $\mathcal{F}$ which can accurately generate the appropriate text response given the multimodal context and the knowledge base as follows,
\begin{equation}
    \mathcal{F}(\mathcal{C}_i, \mathcal{K}|\boldsymbol{\Theta}_F)\rightarrow{\mathcal{R}_i},
     \label{eq1}
 \end{equation}
 where ${\boldsymbol{\Theta}_F}$ represents the model parameters.

\subsection{Dual Knowledge Selection} 
% The knowledge is of crucial importance to the textual response generation in the context of multimodal task-oriented dialog systems. For example, as shown in Figure~\ref{knowledge_example}, the multimodal dialog system can generate the proper response~(\emph{i.e.,} $u_6$) only conditioned on referring to the specific knowledge of the restaurant `Inaniwa Yosuke', such as its domain, location and  representative food in the knowledge base.
To effectively leverage the  entity knowledge, the premise  is to correctly select the related knowledge entities from the whole knowledge base for the given multimodal context.
Considering the multimodal nature of the given context, we devise the dual knowledge selection with  the \emph{\mbox{text-based} knowledge selection} and \emph{vision-based knowledge selection}.
Specially, the text-based and vision-based knowledge selections aim to retrieve the related knowledge entities according to the textual and visual modality of the given context, respectively.
% For simplicity,  we temporally omit the subscript
% $i$ that indexes the training samples.

\textbf{Text-based Knowledge Selection.} 
To capture the related knowledge entities concerning the textual context,
we directly  judge which knowledge entity in the knowledge base is mentioned in the given textual context.
% utilize the names of entities appeared in the textual context to match those in the knowledge base.
% retrieve the knowledge via the entity name.
Namely, for each entity $e_p$ in the knowledge base, 
we check whether it appears in the given textual context. If it appears, we select its attributes $\mathcal{A}_p$ as the related knowledge. 
Notably, we here only consider attributes rather than images due to the fact that the attribute knowledge is essential to understanding user's intentions and generating the text response~\cite{DBLP:conf/mm/ZhangLGLWN21}.
% Notably, considering the semantic supplement roles of knowledge towards the multimodal dialog context, we here only take attributes into consideration rather than corresponding images.
% we take the entity name as the query to retrieve the given textual context. 
% In particular, if the name of  $e_p$ appears in the context,  the textual context knowledge selection module would return the corresponding attributes $\mathcal{A}_p$, and the selection module would return null otherwise.
In this vein, we can obtain the overall knowledge set involved with the textual context, denoted as $\mathcal{K}_t^A =\mathcal{A}_1^t \cup \mathcal{A}_2^t \cup \cdots\cup \mathcal{A}_{N_k^t}^t$, where $\mathcal{A}_m^t$ is the attribute set of the  $m$-th related knowledge entity and $N_k^t$ is the  number of knowledge entities appearing in the textual context.

\textbf{Vision-based  Knowledge Selection.} 
As aforementioned, the goal of the vision-based knowledge selection is to find the related knowledge entities for the given dialog context with its visual modality. As for the same entity, there can be various images characterizing it,  and thus
% Considering the visual diversity of entities~(\emph{e.g.,} images of one scenic spot from different perspectives), 
we employ the visual feature similarity to select the related knowledge for the visual context.
% Regarding the visual context related knowledge filter, 
% ~(\emph{i.e.,} $\mathcal{I}_1, \mathcal{I}_2,\cdots,\mathcal{I}_{N_K}$)
To be specific, we first extract the visual features of entities in the knowledge base $\mathcal{K}$ and images in $\mathcal{V}$ of the given context 
with ViT-B/32~\cite{DBLP:conf/iclr/DosovitskiyB0WZ21} pretrained by CLIP~\cite{DBLP:conf/icml/RadfordKHRGASAM21}, due to its superior performance in various computer vision tasks~\cite{DBLP:journals/corr/abs-2111-08919,DBLP:journals/corr/abs-2203-15334}.
% with the pretrained Residual Network (ResNet)~\cite{DBLP:conf/cvpr/HeZRS16}, due to its superior performance in the computer vision tasks~\cite{Wei2020NeuralMC,DBLP:conf/eccv/LinMBHPRDZ14}. 
Thereafter,  
% similar to[AAA],
% as for each image $v_j$ in $\mathcal{V}$, the visual context-aware knowledge filter performs a latent-space lookup between $v_j$ and each image of $\mathcal{K}$ 
% for the visual knowledge $\mathcal{I}_p$ of each knowledge entity $e_p$
for each image $v_j$ in $\mathcal{V}$, we  measure its similarity to each image of entities in $\mathcal{K}$ 
based on the cosine similarity between their corresponding visual features, and select the top $k$ most similar knowledge entities. Similar to the text-based knowledge selection, we also only consider the semantic knowledge (\emph{i.e.,} attributes) of them.
% analogously returns the corresponding attribute set of the top $k$ most similar knowledge entities. 
In this way, we can acquire the related knowledge set conditioned on the visual context as $\mathcal{K}_v^A=\mathcal{A}^v_1 \cup \mathcal{A}^v_2\cup \cdots \cup \mathcal{A}^v_{N_V}$,  where $\mathcal{A}^v_j$ refers to the related semantic knowledge of the image $v_j$~(\emph{i.e.,} attributes  of the related knowledge entities of image $v_j$).

\subsection{Dual Knowledge-enhanced Context Learning}
To accurately capture the user's intention hidden in the \mbox{multimodal} context, 
% mining the semantic relation between the textual context and the visual context merits our special attention.
% Meanwhile, the knowledge base can effectively promote the response generation via associating the given context with the corresponding knowledge entities.
% Meanwhile, the knowledge base is of crucial importance to the response generation. For example, as shown in Figure~\ref{knowledge_example}, the multimodal dialog system can generate the proper response~(\emph{i.e.,} $u_6$) only conditioned on holding the specific knowledge of `Inaniwa Yosuke', such as the location, domain and the representative food.
% can effectively promote the response generation via associating the given context with the corresponding knowledge entities.
% and gives the specific example of the desired food by the visual context.
% Towards this end,
we design the dual \mbox{knowledge-enhanced}  context learning scheme with two  modules: \emph{knowledge-enhanced context representation} and \emph{dual cross-modal representation refinement}, where the semantic relation between the textual context and the visual context is mined in the latter module. 
For simplicity,  we temporally omit the  subscript
$i$ that indexes the training samples.
\subsubsection{Knowledge-enhanced Context Representation}
% \subsubsection{Knowledge-enhanced Dual Multimodal Learning}

% In a sense, the textual context can indicate the user intention from the  perspective, while the visual context can further reinforce the user intention expression via intuitive images from the local perspective.
% For example, as exhibited in Figure~\ref{knowledge_example}, the user exhibits  the intention of finding a restaurant via the textual context~(\emph{e.g.,} `Japanese food', `udon restaurant' and `ochard road'), and gives the specific example of the desired food by the visual context.
% Meanwhile, the knowledge base can further promote the response generation via associating the given context with the corresponding knowledge entities.
% Therefore, based on the corresponding knowledge of different modality context, we here first conduct the Knowledge-enhanced textual context learning and Knowledge-enhanced visual context learning, respectively.
% As shown in Figure~\ref{knowledge_example}, 
% Having obtained the related knowledge of the given multimodal dialog context, we here proceed to learn the knowledge-enhanced context representation.
% Knowledge-enhanced Context Representation.
In the multimodal dialog, the textual context tends to convey the user's intention from a global perspective, while the visual context would exert roles from the local perspective by reinforcing certain local intention via  intuitive images. 
As shown in Figure~\ref{example_MDialog}, the textual context  generally indicates the user's intention of finding a restaurant and a shopping mall with detailed requirements~(\emph{e.g.,} domain and delivery), while the visual context~(\emph{i.e.,} the image in $u_7$) only  exhibits the desired shopping mall.
Therefore,  we conduct the \emph{global knowledge-enhanced textual representation learning} and \emph{local knowledge-enhanced visual representation learning}.

% In a sense, in the multi-modal dialog, as shown gin Figure~\ref{knowledge_example}, the textual context tends to illustrate the user intention from a global perspective~(\emph{i.e.,} the intention expression of finding a restaurant), while the visual context can enhance the user intention expression from a local manner~(\emph{i.e.,} the specific photo of the desired food).
% Therefore, based on the corresponding knowledge of different modality context, we here first conduct the Knowledge-enhanced textual context learning and Knowledge-enhanced visual context learning, respectively.

% \textbf{Knowledge-enhanced Global Textual Context Learning.} 
\textbf{Global Knowledge-enhanced Textual Representation Learning.} 
Considering the global role of the textual context, 
% in exhibiting the user intention, 
we jointly utilize the related knowledge of both textual and visual context to promote the textual context learning.
% both modality knowledge is essential to be integrated to 
% Accordingly, it is essential to incorporate both the textual context related knowledge and the visual context related knowledge into the textual context.
% which is essential to integrate the whole knowledge to facilitate the response generation.
In particular, we merge the textual context $\mathcal{T}$ and the  related knowledge of both textual and visual context~(\emph{i.e.,} $\mathcal{K}_t^A$ and $\mathcal{K}_v^A$)  as a whole $\mathcal{X}_t=[\mathcal{T}, \mathcal{K}_t^A, \mathcal{K}_v^A]=[x_t^1, x_t^2, \cdots, x_t^{N_t}]$.
Here, $x_t^q$ denotes the $q$-th token and $N_t$ refers to the total number of tokens. 
% considering the prominent performance of BART in modeling the textual information~\cite{DBLP:conf/emnlp/YuDLF21,DBLP:conf/wsdm/GaoZWDC0022}, 
In particular, we first obtain the initial embedding of $\mathcal{X}_t$, denoted as $\mathbf{E}_t\in \mathbb{R}^{{N_t}\times{D}}$, by the position-wise embedding layer of BART in Eqns.~($\ref{eq1_pre}$) and ($\ref{eq2_pre}$).
Thereafter, to capture the semantic representation, we  feed the initial embedding $\mathbf{E}_t$ into the bidirectional encoder $\mathcal{B}_e$ of BART defined in Eqn.~(\ref{eq3_pre})  as follows,
\begin{equation}
    \mathbf{T}_t={\mathcal{B}_e}(\mathbf{E}_t),
     \label{eq6}
 \end{equation}
 where $\mathbf{T}_t\in \mathbb{R}^{{N_t}\times{D}}$ is the knowledge-enhanced representation of the textual context.

\textbf{Local Knowledge-enhanced Visual Representation Learning.}  As aforementioned,  each image of the multimodal context can be associated with certain knowledge entities.
In light of this,  we aim to utilize the corresponding knowledge~(\emph{i.e.,} attributes of the related knowledge entity) to enhance the visual context representation.
{\color{black}
Notably, as for images shown by the agent (\emph{e.g.,} images in $u_4$ of Figure~\ref{example_MDialog}), we directly merge these images into $\mathcal{V}$ as the dialog context.
}

In particular, 
% we adopt ViT-B/32 pretrained by CLIP as the visual encoder $\mathcal{B}_v$.
% , which has delivered the superior performance in various computer vision tasks~\cite{DBLP:journals/corr/abs-2111-08919,DBLP:journals/corr/abs-2203-15334}. 
given the set of images $\mathcal{V}=\{v_1, v_2, \cdots, v_{N_V}\}$, we first employ ViT-B/32 pretrained by CLIP  to encode each image $v_j$ and obtain the visual representation as follows,
 \begin{equation}
    \begin{split}
    \begin{cases}
    % \begin{aligned*}
    \mathbf{v}_j = \mathcal{B}_v({v_j}), j=1,2,\cdots,N_V,\\
    \mathbf{E}_v = [\mathbf{v}_1; \mathbf{v}_2; \cdots; \mathbf{v}_{N_V}]^\top,
    % \end{aligned*}
    \end{cases}
    \end{split}
    \label{eq5}
  \end{equation}
where $\mathbf{E}_v\in \mathbb{R}^{{N_V}\times{D}}$ refers to the initial  representation of the visual context.

Considering heterogeneity between images and their related semantic knowledge, instead of the direct merging operation used in the textual context learning, we 
% each image and its corresponding knowledge caused by distinct data distribution among different modalities,
% we discard the merging operation used in the textual context learning and
resort to the  dot-product attention mechanism~\cite{DBLP:journals/corr/VaswaniSPUJGKP17}, which has been proven to be effective in many multimodal tasks~\cite{DBLP:journals/tois/ChenSRZCN20,DBLP:conf/emnlp/YuDLF21,DBLP:conf/aaai/ChaplotSPRS18,DBLP:journals/tog/FriedTZFSGGJTA19,DBLP:conf/sigir/WenSYZN21}, such as multimodal abstractive summarization~\cite{DBLP:conf/emnlp/YuDLF21}, task-oriented language grounding~\cite{DBLP:conf/aaai/ChaplotSPRS18}, and video editing~\cite{DBLP:journals/tog/FriedTZFSGGJTA19}. 
To be specific, given the related knowledge $\mathcal{A}^v_j$ of the image $v_j$, we first acquire the knowledge embeddings $\mathbf{K}_v^j \in \mathbb{R}^{{N_v^{j}} \times D}$ by the position-wise embedding layer of BART in Eqns.~($\ref{eq1_pre}$) and ($\ref{eq2_pre}$). ${N_v^{j}}$ is the total number of tokens in $\mathcal{A}^v_j$.
Thereafter, we adopt the dot-product attention mechanism  to distinguish informative knowledge tokens towards the representation of $v_j$. Formally, we can obtain the knowledge-enhanced visual representation $\tilde{\mathbf{v}}_j$ of the image $v_j$ as follows,
% we resort to the dot-product attention mechanism which has been proven to be effective in many tasks, such as multimodal abstractive summarization~\cite{DBLP:conf/emnlp/YuDLF21}, task-oriented language grounding~\cite{DBLP:conf/aaai/ChaplotSPRS18} and video editing~\cite{DBLP:journals/tog/FriedTZFSGGJTA19}. 
% In particular, the Knowledge-enhanced representation $\tilde{\mathbf{v}}_j$ of the image $v_j$ can be obtained as follows,
\begin{equation}
    \begin{split}
    \begin{cases}
    \bar{\mathbf{v}}_j = {\mathbf{v}_j^\top}{\mathbf{W}_v^k},\\ %1*D
    \bar{\mathbf{K}}_v^j = {\mathbf{K}_v^j}{\mathbf{W}_k^k},\\ %N*D
    \mathbf{a}_j = softmax(\bar{\mathbf{v}}_j ({\bar{\mathbf{K}}_v^j})^\top),\\ %1*N
    \tilde{\mathbf{v}}_j = LN({\mathbf{v}_j+ (\mathbf{a}_j}{\mathbf{K}_v^j})^\top),
    \end{cases}
    \end{split}
    \label{eq8}
\end{equation}
where $\mathbf{W}_v^k$ and $\mathbf{W}^k_k$ are the to-be-learned
transformation matrices, which aim to project the visual representation~(\emph{i.e.,} $\mathbf{v}_j$) and knowledge embeddings~(\emph{i.e.,} $\mathbf{K}_v^j$) into the same space, and obtain the corresponding latent representations~(\emph{i.e.,} $\bar{\mathbf{v}}_j$ and $\bar{\mathbf{K}}_v^j$). 
$\mathbf{a}_j\in \mathbb{R}^{N_v^{j}}$ is the confidence vector, which denotes different confidence levels of tokens in the knowledge $\mathcal{A}^v_j$ towards the image representation $v_j$. $softmax(\cdot)$ denotes the softmax activation function.  
$LN(\cdot)$ represents the layer normalization operation, which contributes to enhancing the model generalization ability.
Ultimately, we use  $\tilde{\mathbf{E}}_v=[\tilde{\mathbf{v}}_1; \tilde{\mathbf{v}}_2; \cdots; \tilde{\mathbf{v}}_{N_V} ]^\top\in \mathbb{R}^{{N_V}\times{D}}$ to denote the knowledge-enhanced representation of all images in the dialog context.
\subsubsection{Dual Cross-modal Representation Refinement}
% Having obtained the Knowledge-enhanced representation of textual context and visual context, we should proceed to how to seamlessly fuse these representations and thus capture the precise user intention.
% As aforementioned, the textual context can exhibit the user intention from the global perspective, while the visual context can further enhance the user intention expression in a local manner. Undoubtedly, both modality context can convey essential cues regarding the user intention. Towards this end, we devise the refinement component to capture the global-local semantic relation between the
% textual context and visual context, and thus characterize the user intention. To be specific, the component consists of the local vision-oriented refinement and the global text-oriented refinement, respectively.
% convey essential cues concerning the user intention, where they
% As aforementioned, both the textual context and the visual context  can convey essential cues concerning the user intention where they  mutually reinforce each other in a dual manner.
In multimodal dialog systems, 
as both modalities serve to express the same user's intention,
it is promising to learn the context of one modality by
referring to the context of the other modality.
For example, as depicted in Figure~\ref{example_MDialog}, the user exhibits his/her intention of finding a restaurant and a shopping mall with multimodal input, including the  textual description (\emph{e.g.,} `an udon restaurant', `in ochard road', and `a shopping mall near it'), and the image for intuitively showing his/her desired shopping mall.
To  fully leverage the  semantic relation between the textual context and the visual context to enhance  the user intention understanding, we devise the dual \mbox{cross-modal} representation refinement component, with both the \emph{vision-oriented representation refinement}  and the  \emph{\mbox{text-oriented} representation refinement} modules.

\begin{figure}[!t]
    \centering
    \includegraphics[scale=0.68]{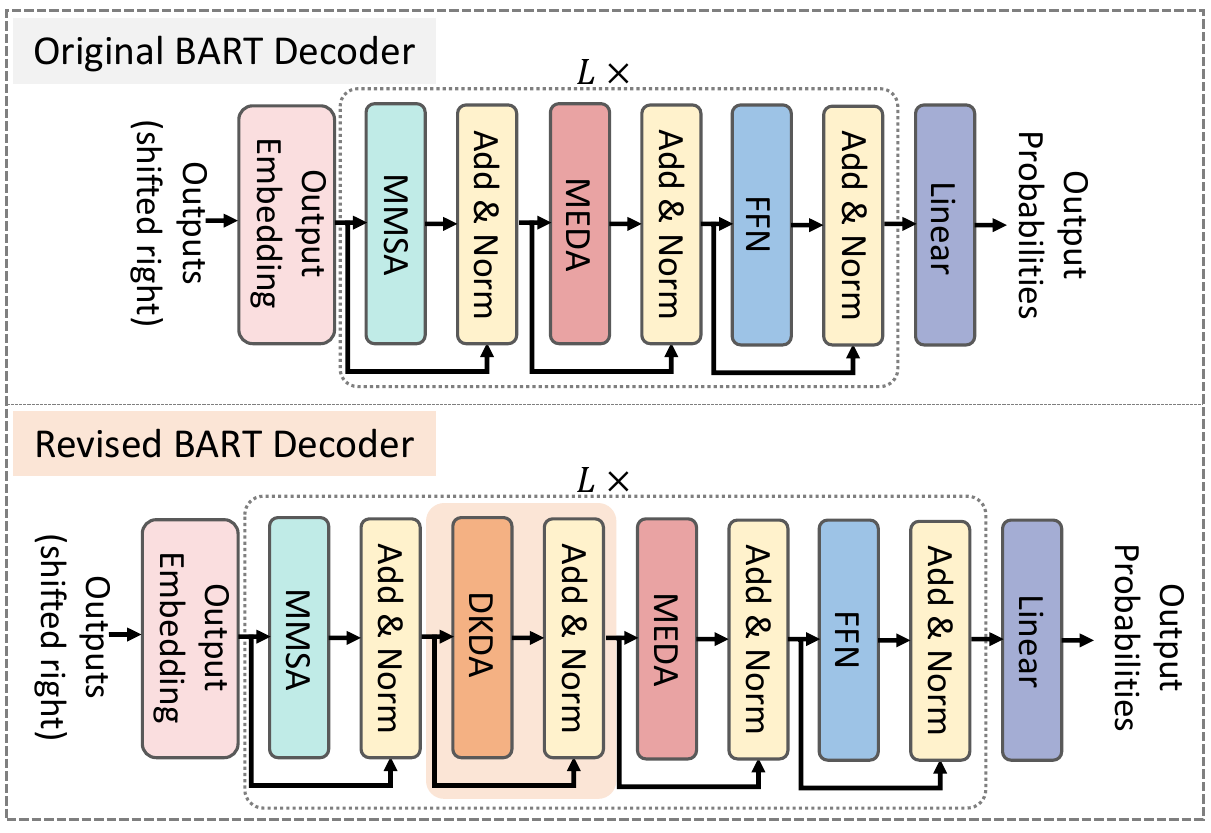}
    \vspace{-1em}
    \caption{{\color{black}Schematic illustration of the original  BART decoder and the revised BART decoder.}}
    \vspace{-1.3em}
    \label{decoder}
\end{figure}

\textbf{Vision-oriented Representation Refinement.} 
In this module, we aim to enhance the visual context representation by referring to the textual modality. Towards this end, we utilize the \mbox{dot-product} attention mechanism to  highlight the informative tokens in the textual context to refine the visual representation. 
% Considering the fact that tokens in the textual context may play different roles to reflect each image $v_j$ in the visual context, 
% we aim to conduct the vision-oriented representation refinement, which can exert informative tokens in the textual context to enhance the representation of images.
% we adopt the dot-product attention mechanism to distinguish the confidences of different tokens.
Specifically, we first obtain the embedding matrix of the textual context $\mathbf{E}_c=[\mathbf{t}_1; \mathbf{t}_2; \cdots; \mathbf{t}_{N_T}]^\top \in \mathbb{R}^{{N_T}\times{D}}$ by the position-wise embedding layer of BART in Eqns.~($\ref{eq1_pre}$) and ($\ref{eq2_pre}$).
% Let $\mathbf{T}_t=[\mathbf{t}_1, \mathbf{t}_2, \cdots, \mathbf{t}_{N_T}]^\top \in \mathbb{R}^{{N_T}\times{D}}$ denote the embedding matrix of the textual context, which can be obtained by the position-wise embedding layer of BART in Eqns.~($\ref{eq1_pre}$) and ($\ref{eq2_pre}$).
$\mathbf{t}_g$ is the embedding of $t_g$, and $N_T$ is the number of tokens in the context.
Then, the vision-oriented representation refinement can be denoted  as follows,
\begin{equation}
    \begin{split}
    \begin{cases}
    % {\mathbf{h}}_j = {\tilde{\mathbf{v}}_j}^\top{\mathbf{W}_v^v},\\ %1*D
    % {\mathbf{U}}_c = {\mathbf{T}_t}{\mathbf{W}_c},\\ %N*D
    \mathbf{o}_j = softmax({\tilde{\mathbf{v}}_j}^\top{\mathbf{W}_v^v} ({\mathbf{E}_c}{\mathbf{W}_c})^\top),\\ %1*N
    \mathbf{P}_j=[[\mathbf{t}_1; {{\mathbf{v}}_j}]; [\mathbf{t}_2; {{\mathbf{v}}_j}]; \cdots; [\mathbf{t}_{N_T}; {{\mathbf{v}}_j}]], & j=1,2,\cdots,N_V,\\
    \hat{\mathbf{v}}_j = {\mathbf{o}_j}{\mathbf{P}_j^\top},
    \end{cases}
    \end{split}
    \label{eq9} 
\end{equation}
where $\mathbf{W}_v^v$ and $\mathbf{W}_c$ are  to-be-learned weight matrices to project different modalities representations into the same semantic space. 
% ${\mathbf{h}}_j\in \mathbb{R}^{1\times D}$ and ${\mathbf{U}}_c\in \mathbb{R}^{N_T\times D}$ are the transferred representation of $v_j$ and textual context, respectively. 
$\mathbf{o}_j\in \mathbb{R}^{{N_T}}$ is the confidence vector to indicate the confidence of tokens in the textual context towards the image $v_j$.
% Inspired by~\cite{DBLP:journals/corr/abs-2104-01122},  $\mathbf{P}_j\in \mathbb{R}^{{2D}\times{N_T}}$ is the initial fusion by directly concentrating the representation of $v_j$ with that of each token.
Inspired by~\cite{DBLP:journals/corr/abs-2104-01122}, instead of simply using the textual embedding vector~(\emph{i.e.,}  $\mathbf{t}_g$) to derive the refined visual representation of $v_j$, we 
 integrate the original visual representation $\mathbf{v}_j$ to 
% concatenate the representation of $v_j$~(\emph{i.e.,}  $\mathbf{v}_j$) with $\mathbf{t}_g$, and thus 
get the 
% revised embedding matrix $\mathbf{P}_j\in \mathbb{R}^{{2D}\times{N_T}}$.
% Inspired by~\cite{DBLP:journals/corr/abs-2104-01122}, 
% to better leverage the textual modality for each image,
% instead of directly using the original embedding matrix~(\emph{i.e.,}  $\mathbf{E}_c$),
% we define $\mathbf{P}_j\in \mathbb{R}^{{2D}\times{N_T}}$ as concatenating  the representation of $v_j$ with that of each token.
% In this way, we can obtain the
enhanced representation 
% of the visual context, $\hat{\mathbf{E}}_v=[\hat{\mathbf{v}}_1; \hat{\mathbf{v}}_2; \cdots; \hat{\mathbf{v}}_{N_V}]\in \mathbb{R}^{{N_V}\times{2D}}$, where $\hat{\mathbf{v}}_j$ is the refined representation
of $v_j$, \emph{i.e,} $\hat{\mathbf{v}}_j$.
Let $\hat{\mathbf{E}}_v=[\hat{\mathbf{v}}_1; \hat{\mathbf{v}}_2; \cdots; \hat{\mathbf{v}}_{N_V}]\in \mathbb{R}^{{N_V}\times{2D}}$ denote the refined visual context representation matrix.

\textbf{Text-oriented Representation Refinement.} 
Analogously, to refine the text context representation learning by referring to the visual modality, we conduct the \mbox{text-oriented} representation refinement.
To be specific, we also employ the dot-product attention mechanism to distinguish informative images in the visual context to refine the textual context.
% obtain the confidence matrix~$\mathbf{S}_E$.
Thereafter, we concatenate the textual context representation $\mathbf{T}_t$ and corresponding distinguished vision representation, and then utilize a fully connected layer to get the final context representation $\mathbf{T}_E \in \mathbb{R}^{N_t \times D}$ as follows,
% Formally, we can obtain the final context representation $\mathbf{T}_E \in \mathbb{R}^{N_t \times D}$ via concatenating the textual context representation $\mathbf{T}_t$ and corresponding distinguished vision representation as follows,
% to enhance the textual representation as follows,
% Moreover, to  distinguish confidences of different images towards the text context representation learning, we devise the  text-oriented representation refinement, which can assign higher confidences to the informative images towards the context.
% To be specific, inspired by~\cite{DBLP:conf/emnlp/YuDLF21}, we also exploit the dot-product attention mechanism as follows,
\begin{equation}
    \begin{split}
    \begin{cases}
    \bar{\mathbf{T}}_t = {\mathbf{T}_t}{\mathbf{W}_t},\\ %N_t*D
    \bar{\mathbf{E}}_v = {\hat{\mathbf{E}}_v}{\mathbf{W}_v^k},\\ %N_V*D
    \mathbf{S}_E = softmax(\bar{\mathbf{T}}_t \bar{\mathbf{E}}_v^\top),\\ %N_t*N_V
    \mathbf{T}_E = [{\mathbf{T}_t}; {\mathbf{S}_E}{\bar{\mathbf{E}}_v}]{\mathbf{W}_f},
    \end{cases}
    \end{split}
    \label{eq10}
\end{equation}
where $\mathbf{W}_t$, $\mathbf{W}_v^k$, and $\mathbf{W}_f$ are the to-be-learned matrices, and $\bar{\mathbf{T}}_t\in \mathbb{R}^{{N_t}\times{D}}$ and $\bar{\mathbf{E}}_v\in \mathbb{R}^{{N_V}\times{D}}$ are the transferred representation of the textual context and visual context, respectively.
$\mathbf{S}_E\in \mathbb{R}^{N_t \times N_V}$ is the confidence matrix, whose $(q,j)$-th entry denotes the confidence of the $j$-th image $v_j$ towards reflecting the $q$-th token $x_t^q$.
% Inspired by~\cite{DBLP:conf/emnlp/YuDLF21}, we obtain the final context representation $\mathbf{T}_E \in \mathbb{R}^{N_t \times D}$ via concatenating the textual context representation $\mathbf{T}_t$ and corresponding distinguished vision representation.
% , and  is the final context representation.

\begin{figure}[!t]
    \centering
    \includegraphics[scale=0.8]{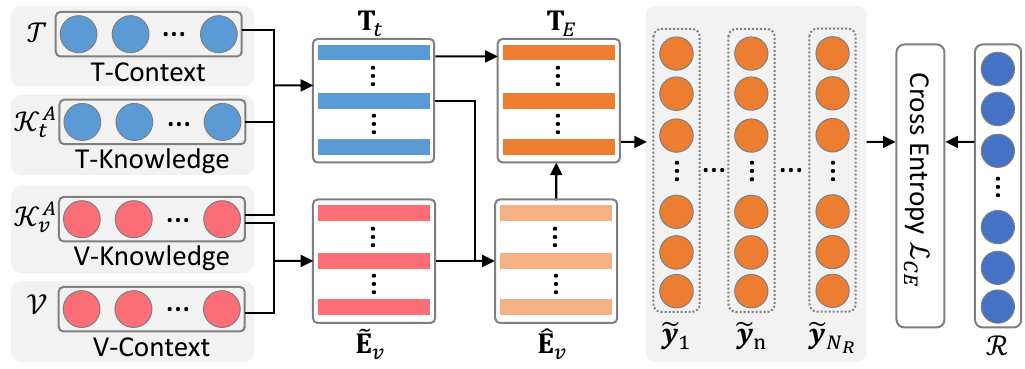}
    \vspace{-0.5em}
    \caption{Workflow of the proposed DKMD. 
    ``T'' and ``V'' denote  Textual and Visual, respectively.}
    \vspace{-1em}
    \label{workflow}
\end{figure}

\subsection{Knowledge-enhanced Response Generation}
% \subsubsection{Knowledge-enhanced Decoder.}
% Having obtained the context representation,
By now, we have obtained the knowledge-enhanced context representation and can move forward to generate the target response.  
In particular, we employ the decoder of BART~(\emph{i.e.,} $\mathcal{B}_d$) in Eqn.~(\ref{eq4_pre}) as our decoder. Although the origin decoder is feasible, the obvious drawback lies in that it  neglects to explicitly exploit the  knowledge for the response generation in the multimodal task-oriented dialog systems.
% to promote the textual response generation performance.
% thus leading to inferior performance. 
% For simplicity, we temporally omit the subscripts $l$ and $n$ that index the decoder layer and the decoding step, respectively.
% For simplicity, we temporally omit the subscript $l$ that indexes the decoder layer and the $n$ that represents the $n$-th decoding step.

% However, the origin decoder neglect the involved knowledge
% In light of this, we revise the original decoder $\mathcal{B}_d$ into  $\tilde{\mathcal{B}}_d$.
% In particular, 
% considering different knowledge tokens may contribute differently in promoting the target response generation,
% % considering different levels  confidence of knowledge tokens towards the response generation,
% we further introduce a dot-product knowledge-decoder attention sub-layer between the masked multi-head self-attention layer and the multi-head encoder-decoder attention layer as follows,
In light of this, as shown in Figure~\ref{decoder}, we revise the original decoder $\mathcal{B}_d$ by introducing a \mbox{dot-product} knowledge-decoder attention~(DKDA) sub-layer, which can distinguish the informative tokens of the context related knowledge and adaptively  utilizes the knowledge to facilitate the text response generation.
% in Eqn.~(\ref{eq4_pre}) 
To be specific, we insert the DKDA sub-layer between the MMSA and the MEDA as follows,
% In light of this, we revise the original decoder $\mathcal{B}_d$ in Eqn.~(\ref{eq4_pre}) by introducing a dot-product knowledge-decoder attention~(DKDA) sub-layer between the MMSA and the MEDA as follows,
%  \begin{equation}
%     \begin{split}
%     \begin{cases}
%     % \begin{align*}
%     \mathbf{q}_l^{S} = LN(MMSA(\mathbf{q}_{l-1}^{dec})+\mathbf{q}_{l-1}^{dec}),\\
%     \mathbf{q}_l^{K} = LN(DKDA(\mathbf{q}_l^{S}, \mathbf{E}_k)+\mathbf{q}_l^{S}),\\
%     \mathbf{q}_l^{E} = LN(MEDA(\mathbf{q}_l^{K}, \mathbf{Z}_L^{enc})+\mathbf{q}_l^{K}), & l=1,2,\cdots,L,\\
%     \mathbf{q}_l^{dec} = LN(FFN(\mathbf{q}_l^{E})+\mathbf{q}_l^{E})
%     % \end{align*}
%     \end{cases}
%     \end{split}
%     \label{eq4_pre_decoder}
%  \end{equation}
\begin{equation}
% \begin{align*}
    \left\{
        \begin{array}{l}
            {\color{black}\mathbf{q}_l^{S} = LN(MMSA(\mathbf{q}_{l-1}^{dec})+\mathbf{q}_{l-1}^{dec}),}\\
            {\color{black}\mathbf{q}_l^{K} = LN(DKDA(\mathbf{q}_l^{S}, \mathbf{E}_k)+\mathbf{q}_l^{S}),}\\
            {\color{black}\mathbf{q}_l^{E} = LN(MEDA(\mathbf{q}_l^{K}, \mathbf{T}_E)+\mathbf{q}_l^{K}),} \\
            {\color{black}\mathbf{q}_l^{dec} = LN(FFN(\mathbf{q}_l^{E})+\mathbf{q}_l^{E}),}
        \end{array}
    {\color{black}\mbox{$l=1,2,\cdots,L,$}}
    \right.
    % \end{align*}
    \label{eq4_pre_decoder}
\end{equation}
where ${{\mathbf{E}}_k}\in \mathbb{R}^{{N_k^d}\times D}$ denotes the embedding of related knowledge $\mathcal{K}_d = \mathcal{K}_t^A\cup \mathcal{K}_v^A$, which can be derived by the position-wise embedding layer of BART in Eqns.~($\ref{eq1_pre}$) and ($\ref{eq2_pre}$). ${N_k^d}$ is the total number of tokens in $\mathcal{K}_d$.
{\color{black}Here, we cast $\mathbf{T}_E$ obtained by Eqn.~($\ref{eq10}$) as the encoder output $\mathbf{Z}_L^{enc}$ in Eqn.~($\ref{eq4_pre}$).}
% Notably, $\mathbf{Z}_L^{enc} = \mathbf{T}_E$, which can be obtained by Eqn.~($\ref{eq10}$).

\begin{algorithm}[!t]
    \caption{Dual Knowledge-enhanced Generative Pretrained Language Model.}
      \label{alg:Framework}
    \begin{flushleft}
    \hspace*{0.02in} {\bf Input:}
    $\mathcal{D}$, $\mathcal{K}$.\\
    % , $\mathbf{A}$, $\lambda$, $\tau$, $\varphi$.\\
    \hspace*{0.02in} {\bf Output:} Predicted distribution $\tilde{\mathbf{y}}$ for all time steps.
      \end{flushleft}
 \begin{algorithmic}[1]
    \State Initialize the parameters $\boldsymbol{\Theta}_F$ based on the generative pretrained language model BART.
    % \State Initialize the latent embedding matrix of personal aspects ${\mathbf{H}^{(0)}}$.\
    \Repeat:
    \State Draw $\mathcal{C}_i$ from $\mathcal{D}$.\
    \State Select the knowledge $\mathcal{K}_t^A$ involved with the textual context in $C_i$ from $\mathcal{K}$.\
    \State Select the knowledge $\mathcal{K}_v^A$ involved with the visual context in $C_i$ from $\mathcal{K}$.\
    \State Compute the global knowledge-enhanced textual  representation $\mathbf{T}_t$ according to Eqn.($\ref{eq6}$).\
    \State Compute the local knowledge-enhanced visual  representation $\tilde{\mathbf{E}}_v$ according to Eqn.($\ref{eq8}$).\
    \State Compute the refined visual context representation $\hat{\mathbf{E}}_v$  according to Eqn.($\ref{eq9}$).\
    \State Compute the refined textual context representation $\mathbf{T}_E$  according to Eqn.($\ref{eq10}$).\
    \State Compute the predicted distribution $\tilde{\mathbf{y}}$  based on the revised decoder of BART.\    
    \State Update $\boldsymbol{\Theta}_F$ according to Eqn.($\ref{eq13}$).\
    \Until: Objective value converges.
 \end{algorithmic}
 \label{algor_all}
%  \vspace{-0.5em}
 \end{algorithm}

Thereinto, 
considering different knowledge tokens may contribute differently in promoting the target response generation,
% considering different levels  confidence of knowledge tokens towards the response generation,
we  define the DKDA sub-layer  as follows,
\begin{equation}
    \begin{split}
    \begin{cases}
    \bar{\mathbf{q}} = {\mathbf{q}^\top}{\mathbf{W}_d},\\ %N_t*D
    \bar{\mathbf{E}}_k = {{\mathbf{E}}_k}{\mathbf{W}_d^k},\\ %N_V*D
    \mathbf{a}_d = softmax(\bar{\mathbf{q}} (\bar{\mathbf{E}}_k)^\top),\\ %N_t*N_V
    \mathbf{q}_k= LN({\mathbf{q}}+({\mathbf{a}_d}{{\mathbf{E}}_k})^\top),
    \end{cases}
    \end{split}
    \label{eq11}
\end{equation}
where for simplicity, we temporally omit the subscripts $l$ and $n$ that index the decoder layer and the decoding step, respectively. $\mathbf{q} \in \mathbb{R}^{{D}}$ refers to the output of the masked multi-head self-attention layer~(i.e., $\mathbf{q}_l^{S}$ in Eqn.($\ref{eq4_pre_decoder}$)).
% ${{\mathbf{E}}_k}\in \mathbb{R}^{{N_k^d}\times D}$ denotes the embedding of related knowledge $\mathcal{K}_d = \mathcal{K}_t^A\cup \mathcal{K}_v^A$, which can be derived by the position-wise embedding layer of BART in Eqns.~($\ref{eq1_pre}$) and ($\ref{eq2_pre}$). ${N_k^d}$ is the total number of tokens in $\mathcal{K}_d$.
$\mathbf{W}_d$ and $\mathbf{W}_d^k$ are  to-be-learned matrices  projecting  $\mathbf{q}$ and  ${\mathbf{E}}_k$ into the same space. 
$\bar{\mathbf{q}} \in \mathbb{R}^{1\times{D}}$ and $\bar{\mathbf{E}}_k \in \mathbb{R}^{{N_k^d}\times{D}}$ are the projected latent representation of $\mathbf{q}$ and ${\mathbf{E}}_k$, respectively.
$\mathbf{a}_d\in \mathbb{R}^{N_k^d}$ is the confidence vector to indicate different levels confidences of  tokens in $\mathcal{K}_d$.
$\mathbf{q}_k$~(\emph{i.e.,} the output of the DKDA sub-layer) denotes the knowledge-enhanced decoder representation.
% Accordingly, we can obtain the knowledge-enhanced decoder representation $\mathbf{q}_k$~(\emph{i.e.,} the output of the dot-product knowledge-decoder attention sub-layer).
% Afterward, we  employ the multi-head encoder-decoder attention layer and the feed-forward network in Eqn.~(\ref{eq4_pre}) to continue the decoding process.

In a nutshell,  for each time step, we can obtain its corresponding predicted distribution $\tilde{\mathbf{y}}$ according to  Eqn.($\ref{pre_predict}$), and thus capture the  predicted token of the current time step  based on the largest element of $\tilde{\mathbf{y}}$.
% the response generation can be represented as,
% \begin{equation}
%     \tilde{\mathbf{y}}_n={\tilde{\mathcal{B}}_d}( {{\mathbf{E}}_k}, \mathbf{T}_E, \tilde{r}_1,\tilde{r}_2,\cdots,\tilde{r}_{n-1}),
%      \label{eq12}
%  \end{equation}
% where $\tilde{\mathbf{y}}_n\in \mathbb{R}^{|\mathcal{U}|}$ is the predicted token distribution for the $n$-th token of the generated response. 
% Furthermore, we can capture the $n$-th token $\tilde{r}_n$ according to the largest element of $\tilde{\mathbf{y}}_n$. 
% Without loss of generality, the decoding process will terminate when the ``End-of-Sequence'' token is generated.
Ultimately, we adopt the cross entropy loss~\cite{DBLP:journals/pr/LiL93} to supervise the response generation as follows,
\begin{equation}
    \mathcal{L}_{CE} = -\frac{1}{N_R} {\sum_{n=1}^{N_R} log(\tilde{\mathbf{y}}_n[t*])},
     \label{eq13}
\end{equation}
where $\tilde{\mathbf{y}}_n[t*]$ refers to the element of $\tilde{\mathbf{y}}_n$ that corresponds to the $n$-th token of the ground truth response $\mathcal{R}$, and $N_R$ is the total number of tokens in $\mathcal{R}$. Notably, the loss is defined for a single sample.
The workflow of DKMD is illustrated in Figure~\ref{workflow}, while the procedure is exhibited in Algorithm~\ref{algor_all}.
% \IEEEraisesectionheading{\section{Model}\label{sec:model}}
\section{Experiment}
In this section, we first introduce the dataset as well as the  experiment setting, and then detail the experiments by answering the following research questions:
\begin{itemize}
    \item \textbf{RQ1}: Does DKMD surpass state-of-the-art methods?
    \item \textbf{RQ2}: How does the knowledge  affect the DKMD?
    \item \textbf{RQ3}: How does the dual cross-modal representation refinement influence the DKMD?
    \item \textbf{RQ4}: {\color{black}How about the sensitivity of DKMD to certain important hyperparameters?}
    %Is DKMD sensitive the location of the encoder layer incorporating the dual \mbox{knowledge-enhanced} context learning? 
    % incorporating
    %  semantic relation hidden in the multimodal dialog
    % \item \textbf{RQ4}: How does AHG-Net perform with  missing data?
    % \item How does the heterogeneous multi-modal information affect the performance of AHG-Net?
\end{itemize}

\begin{table}[!t]
    \centering
  %  \vspace{-1em}
    \caption{Detailed statistics of the MMConv dataset.}
    \begin{tabular}{|l||r|}
    \hline
    Entry                                  & Number         \\ \hline
    \#dialogues                            & $5,106$          \\ \hline
    \#turns                               & $39,759$       \\ \hline
    \#single-modality dialogues             & $751$  \\ \hline
    \#multi-modality dialogues              & $4,355$ \\ \hline  
    \#single-domain dialogues             & $808$  \\ \hline
    \#multi-domain dialogues              & $4,298$ \\ \hline       
    \#entities in the knowledge base  & $1,771$         \\ \hline
    % \#total images                         & $113,953$        \\ \hline
    \end{tabular}
    % \vspace{-2em}
    \label{dataset}
\end{table}

\begin{figure}[!b]
    \centering
    \vspace{-1.5em}
 \subfigure[Training Loss.]{
  \includegraphics[scale=0.56]{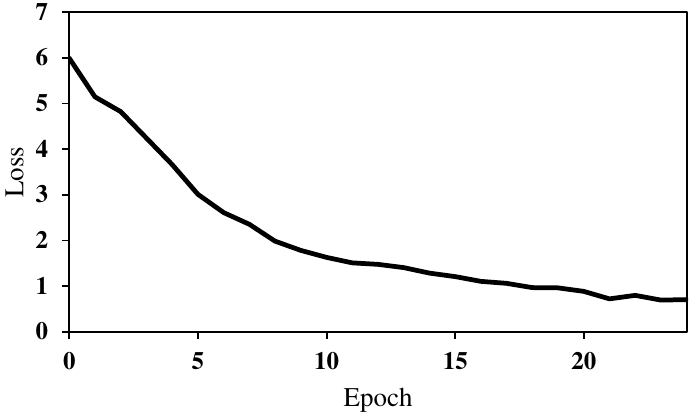}
  %\label{fig:subfig1}
  }
 \subfigure[Testing BLEU-1~($\%$).]{
  \includegraphics[scale=0.56]{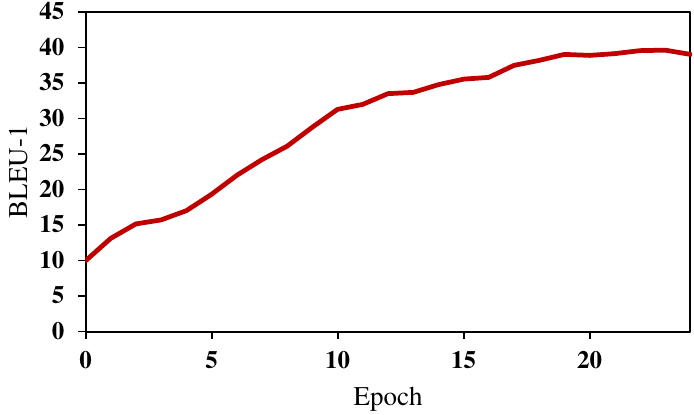}
  %\label{fig:subfig2}
  }
  \vspace{-1em}
  \caption{Convergence analysis of our proposed DKMD.}
    % \vspace{-1em}
    \label{fig_loss}
\end{figure}

\subsection{Dataset}
As a matter of fact, existing efforts  evaluate  their models on the publicly available dataset MMD built by Saha et al.~\cite{DBLP:conf/aaai/SahaKS18} from the fashion domain. 
{\color{black}
However, there is no unique name for each entity in the knowledge base of MMD, which cannot support our model to select the textual context related knowledge based on the entity name. Hence, in this work, we did not choose this dataset for evaluation due to the fact that MMD only allows the knowledge referring by the visual dialog context.
}

Instead, towards comprehensive knowledge referring, we employed the more recently released public dataset MMConv~\cite{DBLP:conf/sigir/LiaoLZHC21}, which is constructed from the general domain and supports knowledge selection from both modalities. 
{\color{black}
The MMConv dataset contains $5,106$ conversations between users and agents spanning five domains: \emph{Food}, \emph{Hotel}, \emph{Nightlife}, \emph{Shopping mall}, and \emph{Sightseeing}. Thereinto, the number of single-modality and multi-modality (\emph{i.e.,} containing both text and image) dialogues in the MMConv dataset are $751$ and $4,355$, respectively. The total number of images in the dataset is $113,953$, while that in dialogues is $10,224$. In addition, there are $6,858$ turns involving images, and the average number of images in each dialogue is $2$.
}
% Thereinto, the MMConv dataset consists of $751$ single-modality conversations and $4,355$ multimodal conversations, where the average turn number are $7.1$ and $7.9$, respectively. 
In addition, the knowledge base of MMConv involves $1,771$ knowledge entities, each of which involves a set of attributes and a few images.
% Similar to existing efforts~\cite{DBLP:conf/mm/LiaoM0HC18,DBLP:conf/mm/NieWHWT19}, we treated every utterance of agents in the conversations as a target response and its former utterance as the context.
More detailed information about the MMConv dataset is summarized in Table~\ref{dataset}.

\subsection{Experiment Setting}
% \subsubsection{Implementation Details.}
We followed the original setting in MMConv~\cite{DBLP:conf/sigir/LiaoLZHC21}, which divides  dialogues into three chunks: $3,500$ for training, $606$ for validation, and $1,000$ for testing.
% Similar to existing efforts~\cite{DBLP:conf/mm/LiaoM0HC18,DBLP:conf/mm/NieWHWT19}, we treated every utterance of agents in the conversations as a target response and its former utterance as the context.
Following the former studies~\cite{DBLP:conf/mm/LiaoM0HC18,DBLP:conf/mm/NieWHWT19}, we treated every utterance of agents in the conversations as a target response and utilized its former two-turn utterances as the given context.
We employed the pretrained BART-large\footnote{https://huggingface.co/facebook/bart-large.} model with 12 layers for encoder and decoder, respectively.
For optimization, we utilized the adaptive moment estimation (Adam) optimizer and set the learning rate as $1e$-$5$.
% {\color{blue} The number of top most similar knowledge entities $k$ was searched in ranges of $[1, 2, 3]$ and set as $1$.}
Moreover, we \mbox{fine-tuned} the proposed DKMD on the basis of the training and validation dataset with $100$ epochs, and reported the performance on the testing dataset.
In addition, we implemented our DKMD  by Pytorch~\cite{DBLP:conf/nips/PaszkeGMLBCKLGA19} and conducted all experiments  on a server equipped with 8 NVIDIA A100 GPUs.
% \subsubsection{Evaluation Metrics.}

Following existing methods~\cite{DBLP:conf/mm/NieWHWT19,DBLP:conf/mm/HeLLCXHY20,DBLP:conf/mm/ZhangLGLWN21}, we 
% selected the following evaluation metrics.
% \textbf{BLEU-\emph{N}.}
adopted BLEU-\emph{N}~\cite{DBLP:conf/acl/PapineniRWZ02} where \emph{N} varies from $1$ to $4$, and Nist~\cite{10.5555/1289189.1289273} as evaluation metrics.
In particular, both BLEU-\emph{N} and Nist can measure the similarity between the generated and target responses, and higher BLEU-\emph{N} and Nist scores denote the more \emph{n}-gram overlap between the generated and target responses~\cite{DBLP:conf/mm/HeLLCXHY20}.

We first experimentally exhibited the convergence of our proposed DKMD. Figure~\ref{fig_loss} demonstrates the changes of the objective loss in Eqn.($\ref{eq13}$) and the testing evaluation metric~(\emph{i.e.,} BLEU-1) with one run of the proposed DKMD. As can be seen, the values first change rapidly and then go steady at last, which effectively corroborates the convergence of our model.

% and higher BLEU scores refer to more n-gram overlaps.
% And based upon BLEU, Nist can dynamically measure the weights of n-grams, where it assigns more weight when the rarer an n-gram is.
\begin{table}[!t]
    \centering
    \caption{Performance comparison among different methods in terms of BLEU-\emph{N}~($\%$) and Nist.}
    % \vspace{-1em}
    \setlength{\tabcolsep}{5.5mm}{
    \begin{tabular}{|l||r|r|r|r|r|}
    \hline
        Models & BLEU-1 & BLEU-2 & BLEU-3 & BLEU-4 & Nist \\ \hline
        MHRED & $15.02$ & $6.66$ & $4.24$ & $2.94$ & $0.9529$ \\ \hline
        % AHRED & $16.38$ & $8.03$ & $5.22$ & $3.75$ & $0.9196$ \\ \hline
        KHRED & $18.29$ & $8.28$ & $4.98$ & $3.36$ & $1.1189$ \\ \hline
        LARCH & $20.86$ & $11.33$ & $7.58$ & $5.58$ & $1.3400$ \\ \hline
        MATE & $30.45$ & $22.06$ & $17.05$ & $13.41$ & $2.3426$ \\ \hline
        UMD & $31.14$ & $21.87$ & $17.12$ & $13.82$ & $2.5290$ \\ \hline
        {\color{black}DialoGPT} & {\color{black}$32.58$} & {\color{black}$23.83$} & {\color{black}$19.22$} & {\color{black}$15.98$} & {\color{black}$2.8182$} \\ \hline
        TREASURE & $34.75$ & $24.82$ & $18.67$ & $14.53$ & $2.4398$ \\ \hline
        {\color{black}T5} & {\color{black}$36.01$} & {\color{black}$27.99$} & {\color{black}$23.38$} & {\color{black}$20.03$} & {\color{black}$3.5890$} \\ \hline        
        \textbf{DKMD} & $\textbf{39.59}$ & $\textbf{31.95}$ & $\textbf{27.26}$ & $\textbf{23.72}$ & $\textbf{4.0004}$ \\ \hline
    \end{tabular}}
    \vspace{-1em}
    \label{rq1}
\end{table}

\subsection{Model Comparison (RQ1)}
To verify the effectiveness of our proposed DKMD, we chose the following state-of-the-art methods on multimodal dialog systems as baselines.
\begin{itemize}
    \item \textbf{MHRED}~\cite{DBLP:conf/aaai/SahaKS18}. 
    This is the first work on the \mbox{task-oriented} multimodal dialog systems, which consists of a hierarchical encoder and a decoder. The hierarchical encoder contains two levels of the gated recurrent units (GRU)~\cite{DBLP:journals/corr/ChungGCB14}, corresponding to encode the utterance and the context, respectively, while the decoder utilizes GRU to generate the response. Notably, this baseline  does not take the knowledge base and the semantic relation hidden in the multimodal context into consideration. % it neglects
    % This is the first work on the multimodal \mbox{task-oriented}  dialog systems, which consists of a hierarchical encoder and a GRU-based decoder.
    % The hierarchical encoder contains two levels of the gated recurrent units~(GRU)~\cite{DBLP:journals/corr/ChungGCB14}, corresponding to encode the utterance and the context, respectively.
    % % , while the decoder utilizes  GRU to generate the response.}
    % % by integrating visual features into the hierarchical recurrent encoder-decoder model~\cite{DBLP:conf/aaai/SerbanSBCP16}.
    % Notably, it neglects the  knowledge base and the semantic relation in the multimodal context.
    % \item \textbf{AHRED}. We devised the baseline based on  MHRED~\cite{DBLP:conf/aaai/SahaKS18}, which introduces the attention mechanism into both encoder and decoder, to distinguish the informative words in one utterance and informative utterances in one context, respectively.
    \item \textbf{KHRED}. Considering the vital role of knowledge in multimodal task-oriented dialog systems, we designed this baseline by incorporating the knowledge into MHRED~\cite{DBLP:conf/aaai/SahaKS18}.
    To be specific, following the knowledge integration way~\cite{DBLP:journals/tip/NieJWWT21}, we utilized the memory network to encode the attribute knowledge, and fed the knowledge representation into the \mbox{GRU-based} decoder to generate text responses.
    % \item \textbf{KHRED}. We designed this baseline by incorporating the knowledge into MHRED~\cite{DBLP:conf/aaai/SahaKS18} by utilizing the memory network to encode the attribute knowledge, and feeding the knowledge representation into the GRU-based decoder to generate textual responses.    
    % which is concatenated with encoded representation and fed into the GRU-based decoder.
    \item \textbf{LARCH}~\cite{DBLP:journals/tip/NieJWWT21} 
    {\mbox{blue}designs a multimodal hierarchical \mbox{graph-based} neural network to model the semantic relation in the given multimodal dialog context, where each word, image, sentence, utterance, dialog pair, and the entire session are treated as nodes. Similar to KHRED, it also integrates the attribute knowledge via a knowledge memory network and adopts the GRU-based decoder for response generation.}
    % designs a hierarchical graph-based neural network to model the semantic relation in the given multimodal context, where each word, image, sentence, utterance, dialog pair, and the entire session are treated as nodes. Similar to KHRED, it also integrates the attribute knowledge via the memory network and adopts the GRU-based decoder for response generation.
    % \item \textbf{LARCH}~\cite{DBLP:journals/tip/NieJWWT21} designs a hierarchical graph-based neural network to model the semantic relation in the given multimodal context, where each word, image, sentence, utterance, dialog pair, and the entire session are treated as nodes. Similar to KHRED, it also integrates the attribute knowledge via the memory network and adopts the GRU-based decoder for response generation.
    % ~\cite{DBLP:conf/mm/LiaoM0HC18} ~\cite{DBLP:conf/mm/ZhangLGLWN21}
    \item \textbf{MATE}~\cite{DBLP:conf/mm/LiaoM0HC18} introduces the Transformer network~\cite{DBLP:journals/corr/VaswaniSPUJGKP17} to capture the context semantic relation~(\emph{i.e.,} semantic dependencies) between the textual context and the visual context, and devises the Transformer-based decoder to generate the text response.
    {\color{black}Although MATE resorts to the original Transformer network, the model parameters are randomly initialized using a Gaussian distribution, which are not pretrained.}
    Notably, it overlooks the knowledge selection  from both the textual and visual perspectives.
    % Notably, it neglects selecting knowledge according to the given context.
    % Notably, it neglects the attribute knowledge.
    
    \item \textbf{UMD}~\cite{DBLP:conf/sigir/CuiWSHXN19} 
    adopts the common multimodal hierarchical encoder-decoder architecture. In particular, it employs a hierarchy-aware tree encoder to learn the taxonomy-guided \mbox{attribute-level} visual representation, and a  multimodal factorized bilinear pooling layer to model the semantic relation between the textual context and visual context. 
    % {\color{blue}  Notably, this method only selects the knowledge according to visual dialog context, overlooking the textual dialog context.}    
    Notably, this method only selects the knowledge according to visual dialog context, overlooking the textual dialog context.
    \item {\color{black}\textbf{DialoGPT }~\cite{DBLP:conf/acl/ZhangSGCBGGLD20} is a pretrained conversational response generation model, which resorts to the Transformer network and is pretrained on large-scale dialogue pairs/sessions extracted from Reddit discussion chains. In particular, the method incorporates the attribute knowledge and is the text-based knowledge-enhanced dialog system.}
    % the only \mbox{textual-based} knowledge-enhanced dialog system.}
     % is a pretrained conversational response generation model, which contains Transformer network and is pretrained on large-scale dialogue pairs/sessions extracted from Reddit discussion chains. In particular, this method is the only \mbox{textual-based} knowledge-enhanced dialog system.}    
    \item \textbf{TREASURE}~\cite{DBLP:conf/mm/ZhangLGLWN21}
    introduces an attribute-enhanced textual encoder, which can enforce the model to adaptively focus on attribute-related keywords and obtain the utterance representation. Besides, this baseline utilizes a sparse graph attention network to learn the context semantic relation and adaptively aggregate the context information. Notably, this method neglects the  knowledge selection from both the textual and visual perspectives.
   \item {\color{black}\textbf{T5}~\cite{DBLP:journals/jmlr/RaffelSRLNMZLL20} is the pretrained language model which incorporates transfer learning techniques. Similar to DialoGPT, this method integrates the attribute knowledge and is also the text-based knowledge-enhanced dialog system.}    
\end{itemize}

Table~\ref{rq1} illustrates the performance comparison among different models with regard to different evaluation metrics.
From this table, we  make the following observations:
\mbox{1) DKMD} consistently outperforms all the baselines across different evaluation metrics, indicating the effectiveness of our proposed DKMD. 
% This may be attributed to the fact that DKMD can effectively leverage the  knowledge from both the text and vision perspectives, and 
This suggests that it is reasonable to incorporate the generative pretrained language model as well as the \mbox{multimodal} context related knowledge in \mbox{multimodal} dialog systems.
2) {\color{black}DKMD surpasses all the baselines that also consider  knowledge (\emph{i.e.,} T5, TREASURE, DialoGPT, UMD, LARCH, and KHRED), which confirms the advantage of our knowledge incorporation manner, \emph{i.e.,} the local-wise and global-wise knowledge enhancement in the multimodal context encoding part, as well as the explicit enhancement in the decoding part. }
% benefit of incorporating the knowledge into the user intention understanding from both the global and local perspectives as well as the explicit response generation.
% 2) DKMD surpasses all the baselines considering knowledge (\emph{i.e.,} TREASURE, UMD, LARCH, and KHRED), which confirms the benefit of integrating the selected dual knowledge into the textual response generation from both the user intention understanding  and the explicit response generation.
% integrating the knowledge into the user intention modeling from both the global and local perspectives.}}
% seamlessly integrates the knowledge into the user intention characterization from both the global and local perspectives. 
% 2) DKMD surpasses TREASURE, UMD, LARCH, and KHRED, which confirms the benefit of exploiting the context related knowledge from both the text and vision perspectives rather than from  only one perspective.
3) MHRED gets the worst performance compared to other methods. This may be attributed to that MHRED not only neglects context related knowledge but also overlooks the semantic relation in the multimodal context, which limits its capability of accurately capturing the user's intention.
4) DKMD, TREASURE, UMD, MATE, and LARCH exceed all the baselines neglecting the context semantic relation (\emph{i.e.,} KHRED and MHRED), which reflects the necessity of exploring the semantic relation hidden in the multimodal dialog context.
{\color{black}5) DKMD achieves a better performance than the baselines that are \mbox{text-based} knowledge-enhanced dialog systems and  also based on pretrained models~(\emph{i.e.,} DialoGPT and T5). The experimental phenomenon  further verifies the necessity and superiority of incorporating visual images and conducting the multimodal dialog systems task. }
\begin{table}[!t]
    % \vspace{-2.5em}
    \centering
    \caption{Ablation study results on knowledge in terms of BLEU-\emph{N}~($\%$) and Nist.}
    % \vspace{-1em}    
    \setlength{\tabcolsep}{4.5mm}{
    \begin{tabular}{|l||r|r|r|r|r|}
    \hline
        Models & BLEU-1 & BLEU-2 & BLEU-3 & BLEU-4 & Nist \\ \hline
        w/o-GlobalK-All & $34.03$ & $25.34$ & $20.61$ & $17.25$ & $3.1755$ \\ 
        w/o-GlobalK-OnlyV & $38.77$ & $30.97$ & $26.25$ & $22.73$ & $3.9112$ \\         
        w/o-LocalK & $38.68$ & $31.13$ & $26.55$ & $23.11$ & $3.8015$ \\
        w-LocalK-AddT & $30.32$ & $22.64$ & $18.51$ & $15.67$ & $3.7723$ \\        
        w/o-DKDA & $38.17$ & $30.56$ & $26.03$ & $22.64$ & $3.8678$ \\ 
        w/o-K-All & $29.99$ & $21.67$ & $17.30$ & $14.23$ & $2.7053$ \\ \hline        
        w/o-TextualK-All & $34.05$ & $26.21$ & $21.78$ & $18.54$ & $3.2022$ \\ 
        w/o-VisualK-All & $38.46$ & $30.98$ & $26.46$ & $23.07$ & $3.8442$ \\ \hline
        \textbf{DKMD} & $\textbf{39.59}$ & $\textbf{31.95}$ & $\textbf{27.26}$ & $\textbf{23.72}$ & $\textbf{4.0004}$ \\ \hline
    \end{tabular}}
    \vspace{-1em}  
    \label{rq2}
\end{table}

\subsection{Effect of Knowledge (RQ2)}
% To thoroughly verify the effectiveness of the knowledge, we compared our DKMD with the following six derivations.
To thoroughly verify the effect of  knowledge in multimodal task-oriented dialog systems, we devised eight derivations as follows.
% to evaluate the specific knowledge module and two derivations to verify the dual knowledge.

% we devised several derivations from two perspectives: the specific module and the knowledge origin.
% To be specific, to validate the 

% We introduced the following derivations to thoroughly verify the effectiveness of the knowledge.
1) \textbf{w/o-GlobalK-All.} To demonstrate the effect of the knowledge in the global text-based learning, we removed all the related knowledge and  only used the textual context as the input of Eqn.($\ref{eq6}$).

2) \textbf{w/o-GlobalK-OnlyV.} To illustrate the necessity of visual context related knowledge in the global text-based learning, we  disabled the visual context related knowledge input of Eqn.($\ref{eq6}$).
% input and kept the textual context and its related knowledge as the input of Eqn.($\ref{eq6}$).

3) \textbf{w/o-LocalK.} To show the benefit of the knowledge in the local vision-based learning, we removed the knowledge refined visual representation obtained by Eqn.($\ref{eq8}$)
% its corresponding knowledge~(\emph{i.e.,} Eqns.($\ref{eq8}$))
and only used the original  visual representation extracted by CLIP~(\emph{i.e.,} Eqn.($\ref{eq5}$)).
% we removed its corresponding knowledge~(\emph{i.e.,} Eqns.($\ref{eq8}$)) and only used the initial  visual representation extracted by CLIP.
% as the input of Eqns.($\ref{eq9}$) and ($\ref{eq10}$).

4) \textbf{w-LocalK-AddT.} To verify  that the textual context related knowledge is redundant in the local vision-based learning, we modified
% concatenated the textual context related knowledge and the visual context related knowledge as
the input of Eqn.($\ref{eq8}$)~(\emph{i.e.,} $\mathbf{K}_v^j$) by  concatenating the textual context related knowledge with the visual context related knowledge.
% rather than only using the visual context related knowledge.

5) \textbf{w/o-DKDA.} To indicate the necessity of injecting the explicit knowledge into the decoder, we discarded the related knowledge~(\emph{i.e.,} the DKDA sub-layer) in the knowledge-enhanced response generation. Namely, we utilized the original decoder of BART~(\emph{i.e.,} $\mathcal{B}_d$). 
% in this variant.

6) \textbf{w/o-K-All.} To verify the importance of the knowledge, we disabled all the knowledge in our proposed DKMD.

7) \textbf{w/o-TextualK-All.} To illustrate the importance of textual context related knowledge in the whole network, we disabled the text-based knowledge and only kept the \mbox{vision-based} knowledge in the global \mbox{text-based} learning~(\emph{i.e.,} removing $\mathcal{K}_t^A$ in $\mathcal{X}_t$ for Eqn.($\ref{eq6}$)) and \mbox{knowledge-enhanced} response generation~(\emph{i.e.,} disabling $\mathcal{K}_t^A$ in $\mathcal{K}_d$ for Eqn.($\ref{eq11}$)).

8) \textbf{w/o-VisualK-All.} To show the roles of visual context related knowledge, we removed the vision-based knowledge in the global text-based learning~(\emph{i.e.,} removing $\mathcal{K}_v^A$ in $\mathcal{X}_t$ for Eqn.($\ref{eq6}$)), local vision-based learning~(\emph{i.e.,} Eqn.($\ref{eq8}$)),  and knowledge-enhanced response generation~(\emph{i.e.,}  disabling $\mathcal{K}_v^A$ in $\mathcal{K}_d$ for Eqn.($\ref{eq11}$)).

% 1) \textbf{DKMD-w/o-GK.} We disabled the related knowledge in the knowledge-enhanced global text-based learning.
% 2) \textbf{DKMD-w/o-LK.} We removed the related knowledge in the knowledge-enhanced local vision-based learning.
% 3) \textbf{DKMD-w/o-DK.} We discarded the related knowledge in the knowledge-enhanced response generation and employed the original decoder of BART.
% 4) \textbf{DKMD-w/o-TK.} We disabled the text-based knowledge and only kept the vision-based knowledge.
% 5) \textbf{DKMD-w/o-VK.} We removed the vision-based knowledge, and only retained the text-based knowledge.
% And 6) \textbf{DKMD-w/o-K.} We disabled all the knowledge  in our proposed DKMD.

\begin{figure*}[!t]
    \centering
    \includegraphics[scale=0.47]{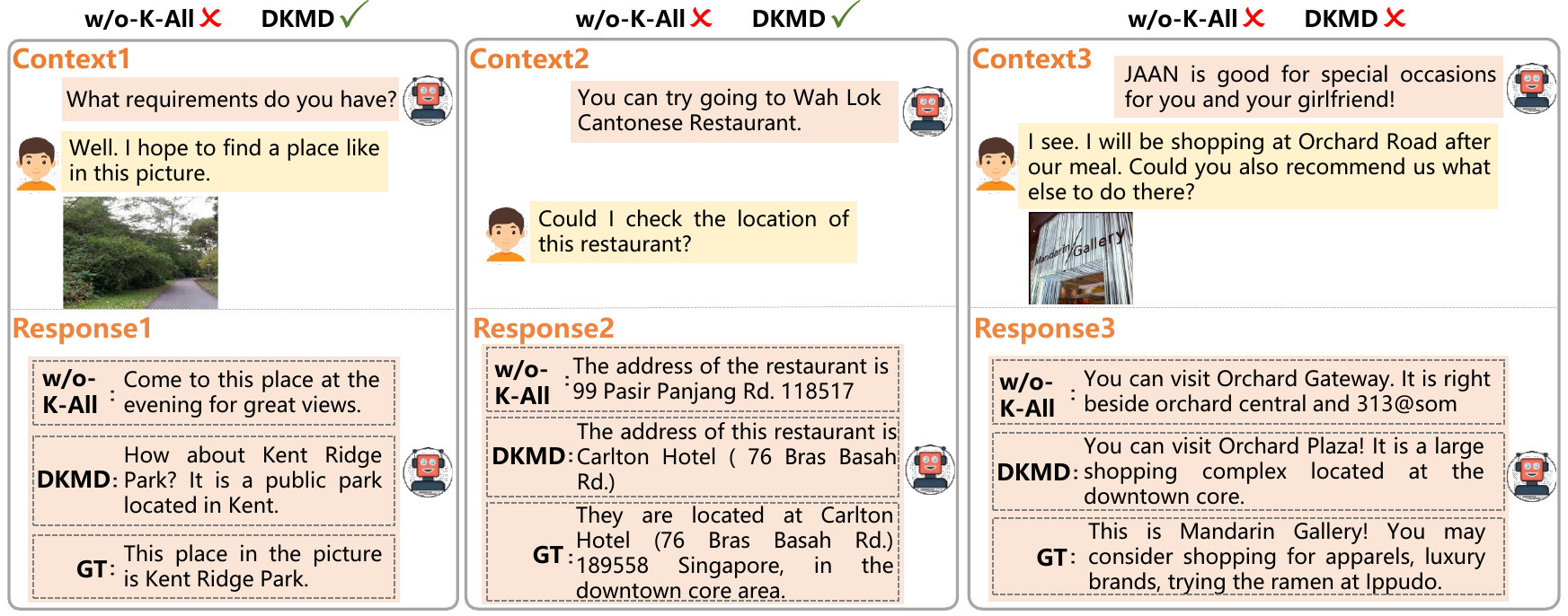}
    % \vspace{-1em}
    \caption{{Comparison between DKMD and w/o-K-All on several testing dialog pairs.  ``GT'' refers to the groundtruth. We represent the correct response of the model with the green tick and the wrong one with the red cross.}}
    \vspace{-1em}
    \label{rq2_case}
\end{figure*}

Table~\ref{rq2} shows the performance of DKMD and its above derivations.
From this table, we have the following observations.
1) DKMD outperforms all w/o-GlobalK-All, w/o-LocalK, \mbox{w/o-DKDA}, and w/o-K-All.
In addition, removing all the knowledge~(\emph{i.e.,} w/o-K-All) leads to the worst performance.
It indicates that disabling the knowledge  anywhere~(\emph{i.e.,} global knowledge-enhanced textual representation learning, local knowledge-enhanced visual representation learning, or \mbox{knowledge-enhanced} response generation) hurts the performance of DKMD.
% and also
% implies that the knowledge  anywhere complement each other and all contribute to the textual response generation in the context of multimodal task-oriented dialog systems.
This may be attributed to that the knowledge  anywhere complement each other and all contribute to the text response generation in the context of multimodal \mbox{task-oriented} dialog systems.
2) w/o-GlocalK-OnlyV performs worse than DKMD.
% DKMD achieves a better performance than w/o-GlocalK-OnlyV.
% and w/o-GlobalK-All, and w/o-GlobalK-All performs worst.
% shows superiority over w/o-GlocalK, 
It demonstrates the effectiveness of the visual context related knowledge in the global text-based learning.
This may be due to that the visual context related knowledge can enhance the global textual context learning and supplement the user intention modeling from the visual perspective, thus boosting the performance of text response generation.
% This may be attributed to that the visual context related knowledge can supplement the textual context related knowledge from the local perspective, and thus boost the user intention modeling.
% DKMD exceeds w-LocalK-AddT
3) w-LocalK-AddT underperforms DKMD, which suggests the redundancy of  considering the textual context related knowledge in the local vision-based learning.  
One plausible explanation is that merging the textual context related knowledge may bring the noise~(\emph{e.g.,} the visual context irrelevant knowledge entities appeared in the textual context) to the visual context learning and thus hurt the performance.
% This may be attributed to that the knowledge in global text-based learning can facilitate the global user intention characterization, while that in local vision-based learning can enhance the local user intention expression.
% This may be attributed to that the knowledge in global text-based learning can facilitate the user intention characterization globally, and that in local vision-based learning can enhance the user intention expression in a local manner.
% 3) We observed that 
% 3) DKMD surpasses DKMD-w/o-TK and DKMD-w/o-VK, which suggests that knowledge of different modalities can 
% 3) We observed that removing any modality of knowledge 
% 4) DKMD surpasses both w/o-TextualK-All and w/o-VisualK-All,
4) Both \mbox{w/o-TextualK-All} and {\mbox{w/o-VisualK-All}}  perform worse than DKMD,
confirming the superiority of considering the knowledge from both the text and vision perspectives.
% in the user intention characterization.
% The underlying philosophy  is that both the textual context and visual context serve to express the same user's intention, and thus their related knowledge also complement each other towards the user intention modeling and response generation.
% , while removing all the knowledge achieves the worst performance.
% This confirms that it is essential to incorporate the knowledge from both the text and vision perspectives.
\mbox{5) w/o-TextualK-All} underperforms  \mbox{w/o-VisualK-All},
% w/o-VisualK-All outperforms w/o-TextualK-All,
which implies that the text-based knowledge contributes more than  vision-based knowledge. 
One possible explanation is that the \mbox{text-based} knowledge derived from the global textual context may be more comprehensive and exert a larger role in facilitating the user intention modeling than the \mbox{vision-based} knowledge. 
% selected from the local visual context.
% information than the vision-based knowledge obtained by the local visual context
% One possible explanation is that the heterogeneity between the visual context and its related semantic knowledge may hinder that the knowledge exerts roles in facilitating the user intention modeling, compared with the textual context and its semantic knowledge.

To gain more deep insights into the influence of knowledge, we showed the comparison between DKMD and w/o-K-All on three testing dialog pairs in Figure~\ref{rq2_case}.
As we can see, DKMD performs  better than w/o-K-All 
in \emph{context1} and \emph{context2} when the knowledge is indispensable to understanding the user's intention.
% in cases when the dialog involves knowledge. 
% For example, we found that the \emph{context1} and \emph{context2} involve knowledge from the text and the vision perspectives, respectively, 
For example, we found that the \emph{context1} and \emph{context2} involve vision-based knowledge and text-based knowledge, respectively, 
and DKMD can generate proper responses and provide accurate information while w/o-K-All fails that.
This phenomenon validates the advantage of incorporating the knowledge into the text response generation in the context of multimodal task-oriented dialog systems.
Nevertheless, DKMD can also yield the failed cases, such as the \emph{context3} in Figure~\ref{rq2_case}.
To be specific, we found that DKMD recommends the wrong shopping mall ``Orchard Plaza'' in the \emph{context3}, which shares similar properties~(\emph{e.g.,} shopping mall, at Orchard Road) with the accurate one ``Mandarin Gallery''.

% \begin{figure}[!b]
%     \centering
%       \vspace{-1em}
%  \subfigure[{\color{blue}The sampled testing multimodal dialog.}]{
%   \includegraphics[scale=0.48]{RQ3-Att-3.pdf}
%   \label{fig:dual_att_subfig1}
%   }
%  \subfigure[{\color{blue}The related text-oriented attention~(\emph{i.e.,} $\mathbf{S}_E$ in Eqn.~($\ref{eq10}$).}]{
%   \includegraphics[scale=0.275]{text2img_1-clip.pdf}
%   \label{fig:dual_att_subfig2}
%   }
%       \vspace{-1em}
%     \caption{Visualization of .} %``Text'', ``Img'' and ``CheckIn'' refer to textual, visual, and check-in posts, respectively. }
%     \label{fig:dual_att}
%     % \vspace{-1em}
% \end{figure}
% Visualization of the learned confidences in the dual cross-modal representation refinement.
\begin{table}[!t]
    \centering
    % \vspace{-1em}
    \caption{Ablation study results on dual cross-modal representation refinement in terms of BLEU-\emph{N}~($\%$) and Nist.} 
    \setlength{\tabcolsep}{4.8mm}{
    \begin{tabular}{|l||r|r|r|r|r|}
    \hline
        Models & BLEU-1 & BLEU-2 & BLEU-3 & BLEU-4 & Nist \\ \hline
        DKMD-w/o-V & $38.25$ & $30.79$ & $26.34$ & $23.00$ & $3.8046$ \\\hline  
        DKMD-w/o-Dual & $38.24$ & $30.17$ & $25.14$ & $21.88$ & $3.8662$ \\ \hline        
        DKMD-w/o-VR & $38.50$ & $30.74$ & $26.09$ & $22.61$ & $3.9252$ \\ \hline
        DKMD-w/o-TR & $38.52$ & $31.05$ & $26.56$ & $23.19$ & $3.8961$ \\ \hline
        \textbf{DKMD} & $\textbf{39.59}$ & $\textbf{31.95}$ & $\textbf{27.26}$ & $\textbf{23.72}$ & $\textbf{4.0004}$ \\ \hline
    \end{tabular}}
    \vspace{-1em}
    \label{rq3}
\end{table}

\subsection{Effect of Dual Cross-modal Representation Refinement (RQ3)}
To explore roles of the dual cross-modal representation refinement, we conduct the comparative experiment with the following derivatives: \textbf{DKMD-w/o-Dual}, \mbox{\textbf{DKMD-w/o-VR}} and \mbox{\textbf{DKMD-w/o-TR}}, where dual cross-modal~(\emph{i.e.,} both vision and text), vision-oriented and \mbox{text-oriented} representation refinement are removed, respectively. In addition, we also introduce the derivative \textbf{DKMD-w/o-V} by removing all the visual information~(\emph{i.e.,} visual context and  knowledge) to verify the necessity of integrating visual information.

Table~\ref{rq3} shows the performance comparison between DKMD and its derivatives.  
% Firstly, as can be seen, DKMD performs better than DKMD-w/o-CR, DKMD-w/o-VR and DKMD-w/o-TR, and DKMD-w/o-CR achieves the worst performance. 
% First, as can be seen, DKMD-w/o-CR~(\emph{i.e.,} removing the dual cross-modal representation refinement) performs worst.
As can be seen, \mbox{DKMD-w/o-Dual}
% ~(\emph{i.e.,} removing the dual cross-modal representation refinement) 
devastates more than DKMD, as compared with DKMD-w/o-VR and DKMD-w/o-TR.
Based on the phenomenon, we had the following two observations. 1) The dual cross-modal representation refinement does benefit the user intention modeling and promotes the text response generation in the context of multimodal \mbox{task-oriented} dialog systems.
% Second, {\color{blue}both DKMD-w/o-VR and DKMD-w/o-TR perform worse than DKMD,}
% DKMD performs better than both DKMD-w/o-VR and DKMD-w/o-TR, 
2) The vision-oriented representation refinement and text-oriented representation refinement  complement each other and both contribute to modeling the  user's intention.
% It suggests that the dual cross-modal representation refinement does benefit the user intention characterization and promote the textual response generation in the context of multimodal dialog systems.
% Second, {\color{blue}both DKMD-w/o-VR and DKMD-w/o-TR perform worse than DKMD,}
% DKMD performs better than both DKMD-w/o-VR and DKMD-w/o-TR, 
% which implies that the vision-oriented representation refinement and text-oriented representation refinement complement each other and both contribute to modeling the  user's intention.
In addition,  we found that \mbox{DKMD-w/o-VR}  underperforms  \mbox{DKMD-w/o-TR}, denoting that the vision-oriented representation refinement contributes more than the text-oriented representation refinement.
% The possible reasons are twofold.1) 
This may be due to that the textual context may convey massive information and play a vital role in delivering the user's intention. 
Therefore, introducing the vision-oriented refinement by referring to the textual modality can significantly improve the visual context understanding.
In contrast, the visual context usually expresses some specific user's intention, like the desired food, thus contributing little to refining the textual context representation and promoting the user's intention learning.
Last but not least, DKMD-w/o-V performs worse than DKMD, denoting
that the visual information plays a pivotal role in the textual response generation of multimodal \mbox{task-oriented} dialog systems.
The underlying philosophy is that the visual information can convey essential clues and thus facilitates the user intention understanding.

% This may be due to  that referring to the textual modality can enhance the understanding of the visual context from the global perspective which contributes more to the user intention, while referring the visual modality only stimulates the user intention expression from the local manner.
% This may be due to the fact that referring the textual modality 
% referring the textual modality can enhance the understanding of the visual context from the global perspective and contributes more to the user intention, while referring the visual modality only stimulates the user intention expression from the local manner which 
% This confirms that the 
% rq3_dual_att

\begin{figure}[!t]
    \centering
    % \vspace{-1em}
 \subfigure[Vision-oriented Representation Refinement.]{
  \includegraphics[scale=0.4]{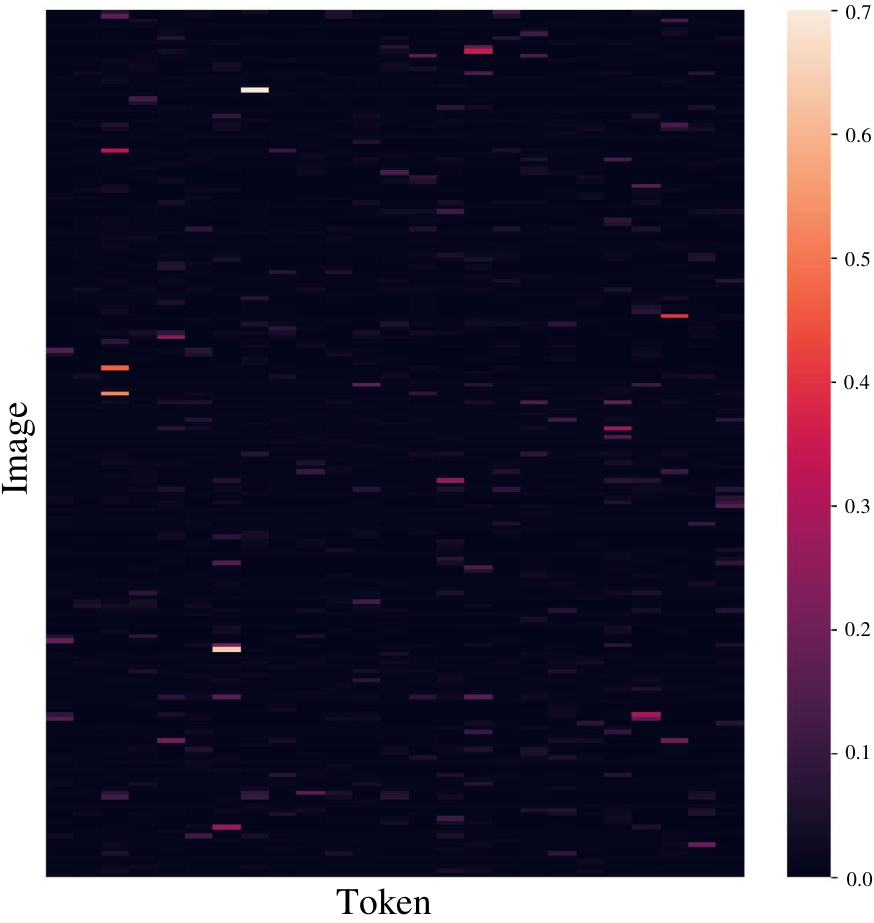}
   \label{att:subfig1}
   }
 \subfigure[Text-oriented Representation Refinement.]{
   \includegraphics[scale=0.4]{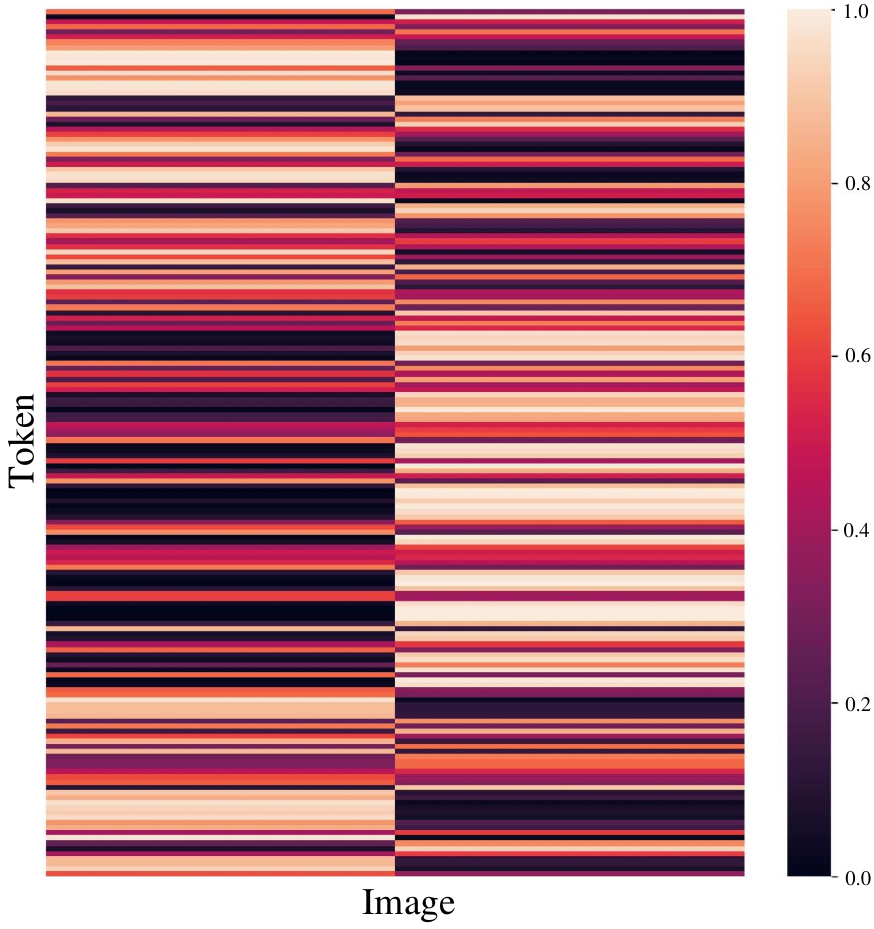}
   \label{att:subfig2}
   }
    \vspace{-1em}
    \caption{Macro illustration of  the dual cross-modal representation refinement.}
    \vspace{-1.5em}
    \label{macro_att}
\end{figure}

\begin{figure}[!b]
    \centering
    \includegraphics[scale=0.43]{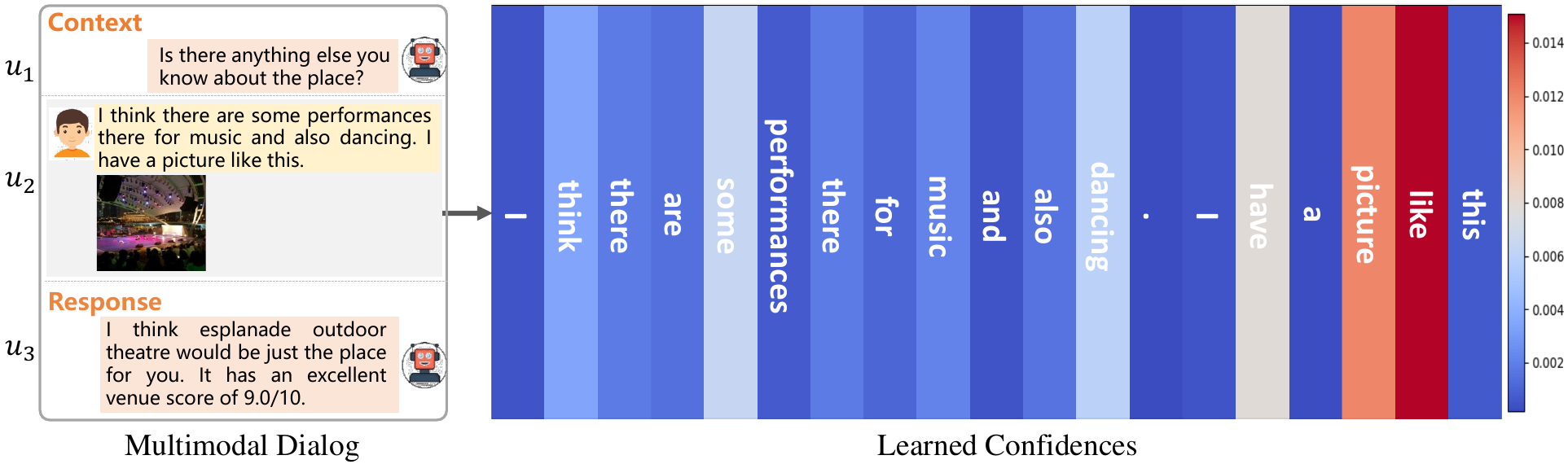}
    \vspace{-1em}
    \caption{Micro illustration of the  learned confidences in the vision-oriented representation refinement concerning the given multimodal dialog.}
    % \vspace{-1em}
    \label{rq3_dual_att}
\end{figure}

Apart from the quantitative comparison between DKMD and its derivatives, we also conducted the macro and micro qualitative illustration about the dual cross-modal representation refinement. 

{\color{black}
% On one hand, we randomly selected several
% testing samples and intuitively illustrated the macro confidences with a thermodynamic diagram in Figure~\ref{macro_att}.
% To be specific, we can extract the confidences~(\emph{i.e.,} $\mathbf{o}_j$ in Eqn.($\ref{eq9}$) and $\mathbf{S}_E$ in Eqn.($\ref{eq10}$)) assigned by the dual cross-modal representation refinement module on the basis of the PyTorch framework.
On one hand, we randomly selected several testing samples and intuitively illustrated the macro confidences with a thermodynamic diagram in Figure~\ref{macro_att}. To be specific, the macro confidences are extracted according to the attention scores (\emph{i.e.,} $\mathbf{o}_j$ in Eqn.($\ref{eq9}$) and $\mathbf{S}_E$ in Eqn.($\ref{eq10}$)) assigned by the dual cross-modal representation refinement module.
}
Each line in Figure~\ref{att:subfig1} corresponds to the token confidences towards each image, while that in Figure~\ref{att:subfig2} denotes the image confidences towards each token. The lighter the color, the higher the
semantic relation between the textual context and the visual
context. As can be seen from Figure~\ref{att:subfig1}, the vision-oriented
representation refinement does assign different levels of
confidences to different tokens for each image, and the similar observation regarding the text-oriented representation refinement can be found in Figure~\ref{att:subfig2}. 
This generally demonstrates that the semantic relation does exist in the multimodal context and both the vision-oriented representation refinement and the text-oriented representation refinement contribute to capturing the semantic relation.
% This demonstrates that the semantic relation does exist in the multimodal context and the dual cross-modal representation refinement can  capture it.

On the other hand, we studied the micro working principle of the dual cross-modal representation refinement.
To be specific, as shown in Figure~\ref{rq3_dual_att}, we randomly selected a testing multimodal dialog and illustrated the learned confidences 
in the vision-oriented representation refinement~(\emph{i.e.,}  $\mathbf{o}_j$ in Eqn.($\ref{eq9}$)) of the utterance $u_2$ with a thermodynamic diagram.
% To gain a better understanding of  the dual \mbox{cross-modal} representation refinement, 
% as shown in Figure~\ref{rq3_dual_att}, we randomly selected a testing
% multimodal dialog and illustrated the learned confidences 
% in the vision-oriented representation refinement~(\emph{i.e.,}  $\mathbf{o}_j$ in Eqn.($\ref{eq9}$)) of the utterance $u_2$ with a thermodynamic diagram.
% intuitively illustrated the learned confidences
% we randomly selected a testing multimodal dialog and illustrated the learned confidences in the dual cross-modal representation refinement with a thermodynamic diagram, as shown in Figure~\ref{rq3_dual_att}. 
% Due to the limited space, we only provided the learned confidences in the text-oriented representation refinement~(\emph{i.e.,}  $\mathbf{S}_E$ in Eqn.($\ref{eq8}$)).
% Each line denotes the  token confidences towards each image. 
The color of the bar denotes the confidence of  tokens towards the image, where the color approaching the orange refers to the larger weight.
% The lighter the color is, the higher the confidence of the image towards the token.
As can be seen from Figure~\ref{rq3_dual_att}, the \mbox{vision-oriented} representation refinement does assign different levels of confidences to different tokens for the given image.
As we can see, our model does identify the informative tokens, such as ``dancing'', ``have'', ``picture'' and  ``like'', and assigns smaller weights to tokens possessing smaller semantic relation with the image~(\emph{e.g.,} ``I'', ``and'',  ``.'', and ``a'').
This suggests that  the dual cross-modal representation refinement can  well capture the semantic relation hidden in the multimodal context.

% where each line corresponds to the token confidences towards each image.
% Similar observation regarding the text-oriented representation refinement can be found in Figure~\ref{fig:dual_att_subfig2}.
% , where each line indicates the image confidences towards each token. 
% This suggests that the semantic relation does exist in the multimodal context and the dual cross-modal representation refinement contributes to  the user intention characterization.

\begin{figure}[!t]
    \centering
    \includegraphics[scale=0.6]{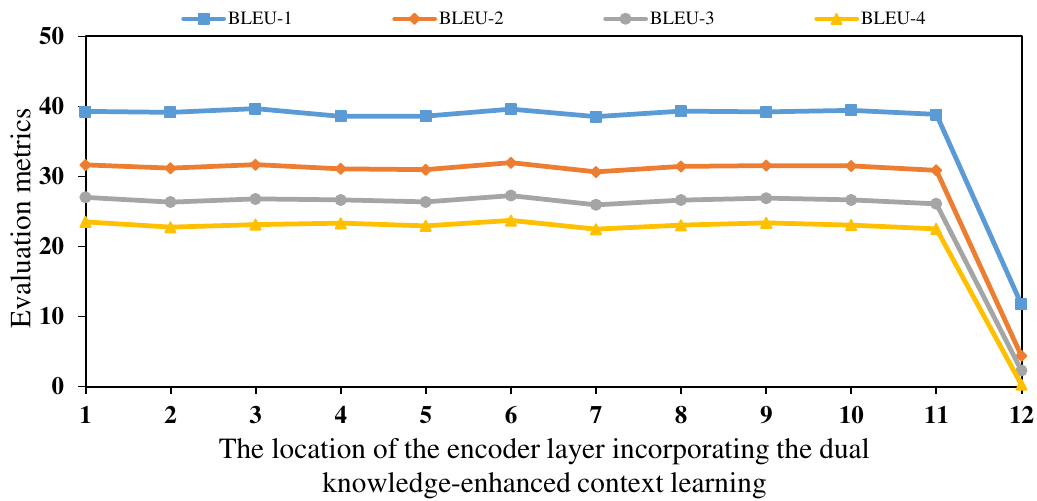}
    % \vspace{-1em}
    \caption{Sensitivity analysis on the location of the encoder layer incorporating the dual knowledge-enhanced  context learning.}
    % \vspace{-2em}
    \label{rq4_sentitive}
\end{figure}
\begin{table}[!b]
    \centering
    \vspace{-1.5em}
    \caption{\color{black}{Performance comparison with different numbers of the selected knowledge entities in our model  in terms of BLEU-$N$ ($\%$) and
Nist.}}
    \setlength{\tabcolsep}{4.2mm}{
    \begin{tabular}{|c||r|r|r|r|r|r|}
    \hline
   Models                 & Number & BLEU-1 & BLEU-2 & BLEU-3 & BLEU-4 & Nist   \\ \hline
    \multirow{3}{*}{DKMD} & 1      & \textbf{39.59}  & \textbf{31.95}  & \textbf{27.26}  & \textbf{23.72}  & 4.0004 \\ \cline{2-7} 
                      & 2      & 39.40  & 31.61  & 26.91  & 23.38  & \textbf{4.0252} \\ \cline{2-7} 
                      & 3      & 39.01  & 31.21  & 26.46  & 22.94  & 3.9023 \\  \hline 
    \end{tabular}}
    \label{entity_num}
\end{table}
% & {\color{red}4}      & 39.01  & 31.21  & 26.46  & 22.94  & 3.9023 \\  \cline{2-7}                     
% & 5      & 39.01  & 31.21  & 26.46  & 22.94  & 3.9023 \\ \hline

\subsection{Sensitivity Analysis}
{\color{black}As there is a stack of layers in the BART encoder, in this part, we performed the sensitivity analysis on the location of the encoder layer incorporating the dual knowledge-enhanced  context learning, and the number of selected knowledge entities.}
% the specific layer incorporating the Dual Knowledge-enhanced Context Learning.

\textbf{Effect of the location of the encoder layer incorporating the dual \mbox{knowledge-enhanced} context learning.} As shown in Figure~\ref{rq4_sentitive}, we enumerated the performance of each layer~(\emph{i.e.,} from $1$-st to $12$-th) in the encoder to perform the dual knowledge-enhanced  context learning.
As can be seen,  our proposed DKMD performs relatively stably when it integrates the dual knowledge-enhanced  context learning at an arbitrary layer between the \mbox{$1$-st} encoder layer to \mbox{$11$-th} encoder layer. 
{\color{black}
Thereinto, our proposed DKMD achieves the optimal performance when it integrates the dual knowledge-enhanced context learning component at the $6$-th encoder layer. One plausible explanation is that integrating the component at the $6$-th layer of the BART encoder can well fuse the learned representation of the component and that of the original BART encoder, and hence fit well with the BART decoder.
}
% Thereinto, our proposed DKMD achieves the optimal performance when it integrates the dual knowledge-enhanced context learning component in the $6$-th encoder layer. One plausible explanation is that integrating the component at the $6$-th layer of the BART encoder can well balance the representation fusion of the component and the original BART encoder, which thus can fit well with the BART decoder.
% Thereinto, our proposed DKMD achieves the optimal performance when it integrates the dual \mbox{knowledge-enhanced}  context learning in the \mbox{$6$-th}  encoder layer. 
% One plausible explanation is that integrating the dual knowledge-enhanced context learning in the \mbox{$6$-th} layer of the BART encoder  can well balance the understanding capability between the knowledge-enhanced multimodal context learning and the original pre-trained BART encoder.
% can well balance the fusion between the representation of textual context  and that of visual context.
{\color{black}
Interestingly, we observed that the proposed DKMD performs the worst when it integrates the dual knowledge-enhanced context learning component at the $12$-th layer. This suggests that the context representation learned by the dual knowledge-enhanced context learning component cannot fit well with the BART decoder, thus leading to the suboptimal performance. This may be due to the fact that there is a semantic gap between the representation learned by the component and that by the original pretrained BART encoder. In contrast, integrating the component at the former layer (i.e., an arbitrary layer between the $1$-st layer to $11$-th layer) allows the fused representation to be further encoded by the subsequent layers in the BART encoder, which may alleviate the semantic gap and achieve a better performance.
}

{\color{black}\textbf{Effect of the number of selected knowledge entities.}
As the number of the selected knowledge entities may affect both the context understanding capability and the decoder process, we also performed the corresponding sensitivity analysis of our model. Table~\ref{entity_num} shows the performance comparison of our model with different numbers of selected knowledge entities. As we can see, our model obtains the optimal performance when we only select the top $1$ most similar knowledge entity. In addition, we noticed that when the number of the selected knowledge entities is three, DKMD achieves the worst performance. One possible explanation is that using the top $3$ most similar knowledge entities may import noise and thus hurt the performance.
% As the number of top most similar knowledge entities may affect both the context understanding capability and the decoder process, we also performed the corresponding sensitivity analysis of our model.
% Table~\ref{entity_num} shows the performance comparison of our model with different number of top most similar knowledge entities. As we can see, our model obtains the optimal performance when we chose top $1$ most similar knowledge entities. In addition, we noticed that
% when the number of top most similar knowledge entities is larger than two, with the increasing number of top most similar knowledge entities, the performance of our proposed DKMD keeps decreasing.
% when the number of top most similar knowledge entities is three, DKMD achieves the worst performance. 
% One possible explanation is that  using top $3$ most similar knowledge entities may import noise and thus hurt the 
% lead to the suboptimal 
% performance. 
}

\section{Conclusion}
In this paper, we tackle the textual response generation task in multimodal task-oriented dialog systems based on GPLMs. 
In particular, we propose a novel dual \mbox{knowledge-enhanced} generative pretrained language model for  multimodal dialog systems,
% knowledge-enhanced dual generative pretrained model for multimodal dialog systems, 
named DKMD, which consists of three pivotal components:  \emph{dual knowledge selection}, \emph{dual knowledge-enhanced context learning}, and \emph{knowledge-enhanced response generation}.
Thereinto, the dual knowledge selection component targets selecting the context related knowledge from the  knowledge base according to both the textual and visual modality of the given context. The dual knowledge-enhanced context learning component aims to integrate the selected knowledge into the multimodal context learning from both global and local perspectives, and thoroughly explore the cross-modal semantic relation. As for the knowledge-enhanced response generation component, we propose a  revised BART decoder with  an additional dot-product knowledge-decoder attention sub-layer. 
Extensive experiments on a public dataset well validate our proposed DKMD and demonstrate the necessity of incorporating the GPLMs and the multimodal context related knowledge in multimodal task-oriented dialog systems.
In addition, we observe that the textual context related knowledge and the visual context related knowledge complement each other and both contribute to the textual response generation. 
Besides, the semantic relation in the multimodal context does exist and should be taken into account. We have released codes and parameters to facilitate other researchers.
% As a residual product, we have released codes and parameters to facilitate other researchers.

Currently, we mainly investigate the potential of GPLMs in the textual response generation task of multimodal task-oriented dialog systems, but ignore the cross-domain relation among different domains.
In the future, we plan to turn to  \mbox{cross-domain} multimodal dialog systems to explore the semantic relation among domains.

\begin{acks}
{\color{black}
This work is supported by the National Key Research and Development Project of New Generation Artificial Intelligence, No.:2018AAA0102502; and the National Natural Science Foundation of China, No.:U1936203;  the Shandong Provincial Natural Science Foundation (No.:ZR2022YQ59).}
\end{acks}

\bibliographystyle{ACM-Reference-Format}
\bibliography{reference}

\appendix

\end{document}